%% file: main.tex
\documentclass[10pt,twocolumn,letterpaper]{article}

\usepackage[pagenumbers]{cvpr} 


\definecolor{cvprblue}{rgb}{0.21,0.49,0.74}
\usepackage[pagebackref,breaklinks,colorlinks,citecolor=cvprblue]{hyperref}
\usepackage{amsmath,amssymb,amsfonts}
\usepackage{algorithmic}
\usepackage{graphicx}
\usepackage{textcomp}
\usepackage{multirow}
\usepackage{times}
\usepackage{epsfig}
\usepackage{bm}
\usepackage{graphicx}
\usepackage{colortbl}
\usepackage{booktabs}
\usepackage{subcaption}
\usepackage{amsmath}
\usepackage{amssymb}
\usepackage{pifont}
\usepackage{url}
\definecolor{tabhighlight}{HTML}{e5e5e5}

\newcommand*{\affaddr}[1]{#1} 

\newcommand{\tablestyle}[2]{\setlength{\tabcolsep}{#1}\renewcommand{\arraystretch}{#2}\centering\footnotesize}
\def\BibTeX{{\rm B\kern-.05em{\sc i\kern-.025em b}\kern-.08em
    T\kern-.1667em\lower.7ex\hbox{E}\kern-.125emX}}
\def\paperID{6616} 
\def\confName{CVPR}
\def\confYear{2025}

\title{BiomedCoOp: Learning to Prompt for Biomedical Vision-Language Models}

\author{%
    \makebox[\linewidth][c]{Taha Koleilat\thanks{Corresponding author}%
    \hspace{3em} Hojat Asgariandehkordi
    \hspace{3em} Hassan Rivaz
    \hspace{3em} Yiming Xiao}
    \\[1ex]
    \affaddr{Concordia University} \\
\hypersetup{urlcolor=black}
        \tt\small \{\href{mailto:taha.koleilat@concordia.ca}{taha.koleilat}, 
        \href{mailto:hojat.asgariandehkordi@concordia.ca}{hojat.asgariandehkordi}, 
        \href{mailto:hassan.rivaz@concordia.ca}{hassan.rivaz}, 
        \href{mailto:yiming.xiao@concordia.ca}{yiming.xiao}\}@concordia.ca
}

\begin{document}
\maketitle
\input{sec/0_abstract}

\input{sec/1_intro}

\input{sec/2_method}
\input{sec/3_results}
\input{sec/4_conclusion}
{
    \small
    \bibliographystyle{ieeenat_fullname}
    \bibliography{main}
}

\input{sec/X_suppl}

\end{document}

%% file: sec/0_abstract.tex
\begin{abstract}

Recent advancements in vision-language models (VLMs), such as CLIP, have demonstrated substantial success in self-supervised representation learning for vision tasks. However, effectively adapting VLMs to downstream applications remains challenging, as their accuracy often depends on time-intensive and expertise-demanding prompt engineering, while full model fine-tuning is costly. This is particularly true for biomedical images, which, unlike natural images, typically suffer from limited annotated datasets, unintuitive image contrasts, and nuanced visual features. Recent prompt learning techniques, such as Context Optimization (CoOp) intend to tackle these issues, but still fall short in generalizability. Meanwhile, explorations in prompt learning for biomedical image analysis are still highly limited. In this work, we propose BiomedCoOp, a novel prompt learning framework that enables efficient adaptation of BiomedCLIP for accurate and highly generalizable few-shot biomedical image classification. Our approach achieves effective prompt context learning by leveraging semantic consistency with average prompt ensembles from Large Language Models (LLMs) and knowledge distillation with a statistics-based prompt selection strategy. We conducted comprehensive validation of our proposed framework on 11 medical datasets across 9 modalities and 10 organs against existing state-of-the-art methods, demonstrating significant improvements in both accuracy and generalizability. The code is publicly available at \tt{\small\href{https://github.com/HealthX-Lab/BiomedCoOp}{https://github.com/HealthX-Lab/BiomedCoOp}}.

\end{abstract}

%% file: sec/1_intro.tex
\section{Introduction}
\label{sec:intro}

 The latest breakthroughs in vision-language models (VLMs) have opened new possibilities for leveraging multi-modal data in diverse applications. Unlike traditional supervised learning that focuses on closed-set visual concepts, models like Contrastive Language-Image Pre-training (CLIP) \cite{radford2021learning}, which align visual and textual information through contrastive pre-training, allow the exploration of open-set visual concepts, thanks to the adoption of natural language supervision. However, the success of these models often relies heavily on the quality of the textual prompts that guide their predictions while full-model fine-tuning for large-scale VLMs is impractical. To mitigate these, prompt learning that optimizes textual prompts in vision-language models \cite{zhou2022learning, zhou2022conditional, khattakMaPLe} has emerged as one of the critical techniques to enhance performance without the need for extensive fine-tuning. Notably, the pioneering work of Context Optimization (CoOp) \cite{zhou2022learning} introduced this approach for CLIP by treating text prompts as learnable context vectors and preserving the pre-trained model weights. Meanwhile, other approaches \cite{gao2024clip, zhang2021tip, huang2024lp++} focus on lightweight few-shot adaptation through Adapters \cite{houlsby2019parameter} and Linear Probes \cite{radford2021learning} to offer parameter-efficient solutions for model adaptation in downstream tasks. 

Different from natural images, biomedical images include a wide range of contrasts and modalities, depending on the image acquisition devices and parameters. These images, such as MRI and ultrasound, often have unique visual appearances that can be more difficult to interpret than typical photographs. In addition, image features (e.g., color, texture, shape, and anatomical context) that are related to physiological and pathological changes are more nuanced and complex to describe, and can differ between image modalities. Finally, due to privacy concerns and the high requirement for clinical expertise, large datasets of well-annotated biomedical images are scarce for developing clinical deep learning models. While VLMs and the associated prompt learning techniques have shown success across natural image datasets and benchmarks, their application in the biomedical imaging domain (e.g., diagnosis), which has distinct challenges, remains largely under-explored.

Due to the unique domain knowledge of biomedical images, the backbone vision-language model for prompt learning may require tailored pre-training for the best outcome. Biomed-specific VLMs, such as BiomedCLIP \cite{biomedclip}—pre-trained on 15 million biomedical image-text pairs from internet resources—are better suited for biomedical tasks \cite{zhao2023clip}. Although few recent works \cite{cao2024domain, fang2024aligning, bie2024xcoop} have investigated prompt learning for biomedical image classification by using natural-image-trained CLIP backbones, their explorations remain in a narrow range of tasks with larger datasets (e.g., chest X-ray and dermatology), which may not be common for other clinical tasks. Notably, these approaches either employ the full training dataset \cite{bie2024xcoop, fang2024aligning} or domain-controlled prompt learning via an additional foundation model, i.e., MedSAM \cite{ma2024segment}, reducing their computational efficiency. In addition, further improvements in accuracy and generalizability to unseen classes are still needed over the existing prompt learning approaches. Therefore, it is highly instrumental and desirable to explore more efficient and robust novel prompt learning techniques based on biomed-specific VLMs (e.g., BiomedCLIP) and conduct large-scale validation on diverse benchmarking datasets of biomedical images. 

In this work, we present an innovative prompt-learning framework based on CoOp \cite{zhou2022learning}, called BiomedCoOp, to facilitate efficient adaptation of CLIP-like VLMs, such as BiomedCLIP, in few-shot biomedical image classification for differential diagnosis. Instead of full model fine-tuning, our approach exclusively focuses on effective strategies for textual prompt optimization with the help of large language models (LLMs), which not only reduce computational overhead, but also preserve the pre-trained model’s foundational knowledge. Specifically, with high disparity across imaging modalities (e.g., ultrasound vs. MRI), we hypothesize that a few-shot learning strategy will benefit the application while addressing data limitations in the biomedical domain. Furthermore, LLMs can help mitigate limitations in human-designed prompt templates in context learning. Our four key contributions include:

\begin{enumerate} 
\item We introduce a novel prompt-learning approach that enhances context vector learning by enforcing semantic consistency through prompt ensembles derived from large language models (i.e., GPT-4) and an effective knowledge distillation strategy.

\item To address the challenge of outlier prompts from LLMs during context learning, which can lead to over-specialization and hinder generalization, we employ a statistics-based pruning strategy to mitigate the risk of ``forgetting" essential biomedical patterns while maintaining sensitivity to diverse disease presentations. 

\item Our study adopts BiomedCLIP \cite{biomedclip} for prompt learning for the first time, and we demonstrate the benefits against general knowledge CLIP in downstream clinical tasks. 

\item We conduct a comprehensive evaluation of our proposed method against existing CLIP prompt learning techniques using 11 diverse biomedical image classification datasets across 9 modalities and 10 organs in few-shot and base-to-novel generalization benchmarks. Our results highlight BiomedCoOp's superior generalizability and robustness across a wide range of medical conditions and imaging modalities.

\end{enumerate}

\section{Related Work}
\label{sec:related-work}
\subsection{Vision-Language Models}
Vision-language models like CLIP \cite{radford2021learning} and ALIGN \cite{jia2021scaling} have transformed multi-modal learning with self-supervised visual and textual representations in a common feature space, enabling remarkable performance in various tasks, including zero-shot classification and cross-modal retrieval. Recent extensions in the biomedical domain include BioViL \cite{boecking2022making}, PubMedCLIP \cite{eslami2021doesclipbenefitvisual}, and BiomedCLIP \cite{biomedclip}, which adapt VLMs to biomedical data using millions of biomedical image-text pairs. However, while effective, these models still require additional task-specific adaptation to capture the fine-grained, disease-specific nuances crucial in clinical applications for better performance \cite{xu2024advances, zhao2023clip}. Such limitations highlight the need for methods that conduct further domain-specific adaptations for target clinical tasks.
\begin{figure*}[!ht]
\centering
{\includegraphics[width=\textwidth]{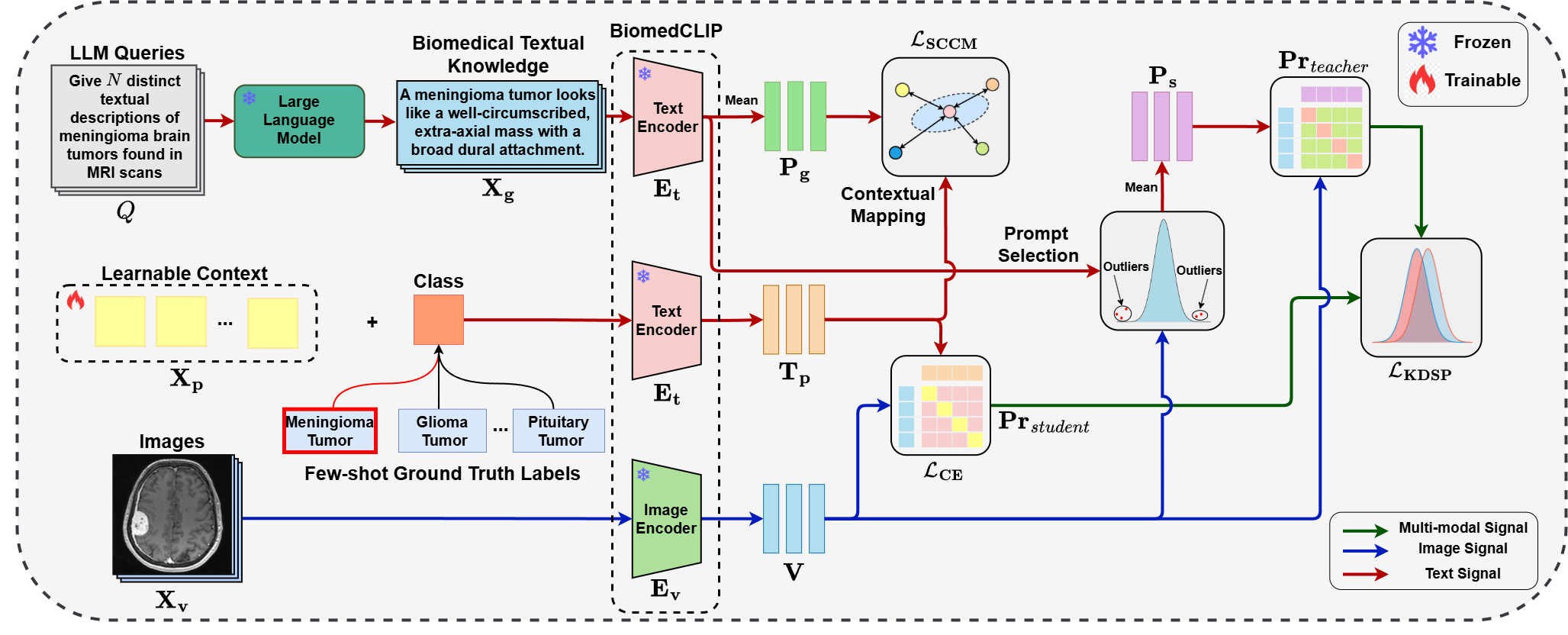}}\vspace{-0.5em}
\caption{
Overview of the BiomedCoOp framework, which combines LLM queries, learnable context tokens, and BiomedCLIP to generate multi-modal representations for biomedical tasks. The method integrates text and image features using prompt ensembling strategies, minimizes cross-entropy and semantic differences, and aligns teacher-student logits, enabling effective few-shot learning for novel biomedical categories.
}
\label{fig:main_figure}
\end{figure*}
\subsection{Prompt Learning}
Prompt learning has emerged as a powerful alternative to traditional model fine-tuning for VLMs, with methods like CoOp \cite{zhou2022learning} and CoCoOp \cite{zhou2022conditional} treating text prompts as the primary learnable component to steer VLMs toward specific tasks. In biomedical applications, prompt learning has shown preliminary potential for few-shot adaptation without altering pre-trained weights \cite{du2024cleft}. Recent methods in the natural vision domain like MaPLe \cite{khattak2023maple} adapt both the vision and language components of CLIP while PromptSRC \cite{khattak2023self} applies self-regulation techniques to improve prompt generalization without sacrificing prior knowledge. Complementing these approaches, KgCoOp \cite{kgcoop} and ProGrad \cite{prograd} refine textual prompts using knowledge and gradient guidance, respectively, to enhance model generalizability. Additionally, ProText \cite{khattak2024learning} stands out with deep prompt learning solely from text data derived from LLMs, enabling cross-dataset and cross-class transfer without requiring labeled images. However, these methods may not be well suited for biomedical applications due to the aforementioned challenges, especially cross-modal differences in anatomical and pathological feature descriptions. Recently, biomed-specific approaches, including ViP \cite{fang2024aligning} and XCoOp \cite{bie2024xcoop}, have enhanced the adaptability of VLMs in clinical tasks by integrating disease-specific terminology and contextual tokens. Domain-Controlled Prompt Learning (DCPL) \cite{cao2024domain} further extends this adaptability by integrating domain-specific biases for both vision and language branches, particularly useful in specialized fields such as remote sensing and medical imaging, where natural-domain prompts fall short. Despite more tailored solutions for biomedical tasks, these methods often require heavier adaptation of both vision and language components and clinical expert intervention, thus potentially limiting their performance, ease of implementation, and generalizability to wider tasks. As suggested by Khattak et al. \cite{khattak2024learning}, robust mapping between learnable context and knowledge from LLMs can benefit the performance and adaptability of VLMs. With human-level performance and stability upgrades of recent LLMs like GPT-4 in diagnostic reports \cite{krishna2024evaluation, gertz2024potential}, integration of the latest LLMs, in addition to biomed-specific VLMs (e.g., BiomedCLIP) may offer an alternative avenue to more conveniently incorporate biomedical domain expertise in VLMs for a data-efficient prompt learning solution, suitable for diverse clinical needs. Yet, this remains to be explored, and we intend to take advantage of these advances in our methods.

\subsection{Few-shot Adaptation of VLMs}
Few-shot adaptation techniques for vision-language models enable task specialization with minimal labeled data to balance generalization and domain specificity. Aside from few-shot-based prompt learning, adapter-based methods like CLIP-Adapter \cite{gao2024clip}, and Tip-Adapter \cite{zhang2021tip} incorporate lightweight modules that adjust visual features while preserving zero-shot capabilities. Specifically, CLIP-Adapter modifies visual embeddings via a compact MLP. Tip-Adapter uses a similarity-based mechanism that blends visual features from the support set directly into the model’s predictor to enhance accuracy with minimal data, but requires careful tuning. Enhanced linear probing methods, such as LP++ \cite{huang2024lp++}, extend adaptation by mixing visual and text features and using data-informed implicit learning rates to achieve competitive few-shot performance without extensive hyperparameter optimization. Finally, CLAP \cite{silva2024closer} further constrains adaptation to stay close to original zero-shot prototypes by using adaptive penalties to prevent overfitting. These categories of methods refine visual embeddings typically at the final layers of VLMs, focusing on adjusting model features. In comparison, prompt learning approaches that target optimizing textual prompt inputs may be more advantageous in computational efficiency and adaptation to unseen classes, particularly in biomedical imaging domains.

%% file: sec/2_method.tex
\section{Methodology}
\label{sec:methods}

An overview of the proposed BiomedCoOp framework is illustrated in Fig.~\ref{fig:main_figure}. By leveraging the BiomedCLIP backbone to encode the rich image and text features, our method proposes two major components to enable effective prompt context learning jointly. Specifically, while the \textit{Semantic Consistency by Contextual Mapping} (SCCM) component aligns text embeddings with general biomedical knowledge via minimizing their distance to class-specific prompts generated by an LLM, we design the \textit{Knowledge Distillation with Selective Prompting} (KDSP) component to refine context mapping through statistics-based prompt selection. The unified learning objective integrates cross-entropy loss, MSE loss of \textit{SCCM}, and Kullback-Leibler (KL)-divergence loss of \textit{KDSP} to ensure accurate and robust model representations.

\subsection{Contrastive Language-Image Pre-training}
CLIP utilizes a vision encoder $\bm{E_v}$ and a text encoder $\bm{E_t}$ to take batches of pre-processed images $\bm{X_{v}}$ $\in \mathbb{R}^{B \times 3 \times H \times W}$ and texts $\bm{X_{t}}$ $\in$ $\mathbb{R}^{B \times L}$ as inputs. Here, $B$ is the number of image/text pairs in a single batch, $H$ and $W$ are the height and width of the images, respectively, and $L$ is the maximum sequence length of the tokenized text input. Specifically, the vision and text encoder networks produce the corresponding feature vectors $\mathbf{V}$ = $\bm{E_v}(\bm{X_{v}})$ and $\mathbf{T}$ = $\bm{E_t}(\bm{X_{t}})$, where $\mathbf{V}$ $\in \mathbb{R}^{B \times D}$,$\mathbf{T}$ $\in \mathbb{R}^{B \times C \times D}$, $C$ is the number of classes and $D$ is the embedding dimension. Finally, the multi-modal feature vectors are aligned in the latent space based on their similarity scores through a contrastive learning objective. The pre-trained CLIP model with rich multi-modal representation has the capacity for zero-shot classification. Given $C$ possible classes, a new image can be classified by computing its similarity to a set of $C$ different text prompts as ``\texttt{a photo of a [CLASS]}", where ``\texttt{a photo of a}" represents the pre-defined context of the textual template while \texttt{[CLASS]} is the class name token. Notably, in CoOp, the context (e.g., ``\texttt{a photo of a}") is learnable and optimized according to a categorical objective. Specifically, the zero-shot inference is made by comparing each text feature $\bm{T}^{(i)}$ of the different text prompts with the image embeddings $\bm{V}$ to get a prediction probability:
\begin{equation} \label{eq:prob_coop}
p(\bm{Y}=i|\bm{V},\bm{T}^{(i)}) = \frac{\exp( \cos(\bm{T}^{(i)}, \bm{V}) /\tau)}{\sum_{j=1}^C \exp( \cos(\bm{T}^{(j)}, \bm{V}) /\tau)},
\end{equation}
where $\tau$ denotes the learnable temperature parameter and $\cos(\cdot, \cdot)$ denotes cosine similarity. The class of the image is determined by taking:
\begin{equation}
    \text{arg}\max_{\bm{i}}p(\bm{Y}=i|\bm{V},\bm{T}^{(i)})
\end{equation}

\subsection{LLM Prompt Ensembling}
While prompt ensembling is shown to benefit prompt learning by bringing diverse text representations \cite{zhang2024prefer}, human-designed prompt templates that were previously adopted in natural vision tasks \cite{liu2023chatgpt} can face obstacles for biomedical domains in sourcing relevant clinical knowledge and sufficient description diversity. Notably, recent studies \cite{krishna2024evaluation, gertz2024potential} on GPT-4 \cite{achiam2023gpt} have validated its performance in tasks related to clinical case reports. Thus, we decided to leverage GPT-4 for LLM-generated prompt ensembling with diverse textual descriptions of class-specific lesions and abnormalities in diagnostic scans. This approach ensures that the contextual structure of the learned prompts in BiomedCoOp reflects the required domain knowledge to help effectively capture essential semantic features. Specifically, for $C$ different classes in a dataset, we generate text prompts $\bm{X_{g}}$ $\in \mathbb{R}^{N \times C \times L}$ comprising $N$ different outputs for each class from text query $Q$:
``\texttt{Give $N$ textual descriptions of visual discriminative features for distinct medical cases of [CLASS] found in [MODALITY]}." 
Unique for biomedical images, we specifically mention the imaging modality in $Q$ as certain classes might overlap in different modalities. $\bm{X_g}$ from the LLM are then encoded to $\bm{T_g}$ = $\bm{E_t}(\bm{X_{g}})$ $\in \mathbb{R}^{N \times C \times D}$. For the SCCM component, all $N$ text embeddings for each class are ensembled by taking the mean to get $\bm{P_g}$ $\in \mathbb{R}^{C \times D}$:
\begin{equation}
    \bm{P_g}^{(c)} = \frac{1}{N}\sum\limits_{i=1}^N\bm{T}_\text{\textbf{g},i}^{(c)}
    \label{eq:general_mean_prompts}
\end{equation}

\subsection{Selective Prompting via Outlier Exclusion}
In the KDSP component, we further refine the learned context by also considering the probability distribution of the LLM-generated prompts. As overly specific prompts can risk overfitting the model on particular disease states and low-scoring prompts that do not align well with relevant biomedical features can hurt accuracy, we propose selective prompt ensembling with outlier pruning. By removing these outliers, the distribution of prompts is refined, ensuring that the selected textual prompt distribution reflects broader biomedical insights, helping the model avoid ``forgetting" key general knowledge while preserving the versatility of the BiomedCLIP model. This enables the model to remain sensitive to both typical and atypical features across diverse disease presentations in biomedical images.

Given a batch of $B$ images, each representing a unique case for a disease class, we first encode these images using the vision encoder $\bm{E_v}$ to obtain image features $\bm{V}$. For each image embedding in $\bm{V}$, we compute its similarity with the corresponding prompt embedding in $\bm{T_g}$.  The similarity measure, typically cosine similarity, helps identify the prompts most relevant to the specific characteristics of the image. The scores $\bm{S}$ are calculated for each prompt by taking the mean of the maximum similarity logits across all image embeddings:

\begin{equation}
    \bm{S} = \frac{1}{B} \sum_{i=1}^{B} \max_{j} (\beta \cdot \bm{T_g}^{(j)} \cdot \bm{V}^{\intercal}),
\end{equation}
where $\beta$ is a scaling factor applied to the logits.

To detect and handle anomalous prompts that deviate from the general distribution, we apply an outlier detection approach using the \textit{Median Absolute Deviation} test statistic. Specifically, we calculate the median $M_{s}$ of the prompt scores $\bm{S}$ and the median absolute deviation $D$:
\begin{equation}
    M_{s} = \mathrm{median}(\bm{S})
\end{equation}
\begin{equation}
    D = \mathrm{median}(|\bm{S} - M_{s}|)
\end{equation}
\\
For a given prompt, we compute the modified $z$-score:
\begin{equation}
    z = \frac{\bm{S} - M_{s}}{D}.
\end{equation}
We select only the $N_s$ prompts that are associated with the modified $z$-scores with absolute values that are lower than a selection threshold $\zeta_s$. Following a similar approach to Eq.~\ref{eq:general_mean_prompts}, we obtain an average prompt encoding $\bm{P_{s}}$ based on the selections.

\subsection{Overall Learning Objective}
The overall learning objective of our few-shot prompt learning framework entails a cross-entropy loss for classification accuracy and the context-mapping-related losses from the \textit{Semantic Consistency by Contextual Mapping} and \textit{Knowledge Distillation with Selective Prompting} components. Their details are described below:
\vspace{0.1cm}
\\
\noindent
\textbf{First}, we aim to learn prompts \( \bm{X_p} \) by optimizing the overall learning objective given \( K \) few-shot ground truth examples for each class. The objective includes the cross-entropy loss of the image-text logits, which is defined as:

\begin{equation}
    \mathcal{L_{\texttt{CE}}} = -\sum_{i=1}^{C} \bm{Y}^{(i)} \log p(\bm{Y}=i|\bm{V}, \bm{T_p}^{(i)}),
\end{equation}

\noindent where \( \bm{Y}^{(i)}\) is the ground truth label for class \( i \), and \( p(\bm{Y}=i|\bm{V}, \bm{T_p}^{(i)})\) is the predicted probability of class \( i \) given images $\bm{V}$ and encoded learnable text prompt $\bm{T_p}^{(i)}$.
\vspace{0.1cm}
\\
\textbf{Second}, since the context to be learned is unified among all the classes, we minimize the difference between $\bm{P_g}$ and $\bm{T}_p$ in the SCCM component to ensure that the general biomedical knowledge is properly learned:
\begin{equation}
    \mathcal{L_{\texttt{SCCM}}} = \sum_{i=1}^{C}||\bm{T_p}^{(i)} - \bm{P_g}^{(i)}||_{2}^{2}
\end{equation}
\vspace{0.1cm}
\\
\noindent
\textbf{Lastly}, to align the distribution of the logits from image embeddings with learnable context prompts (student logits) and the logits from image embeddings with selective LLM-generated text embeddings (teacher logits), we minimize the KL divergence between these two distributions in the KDSP component:
\begin{equation}
    \mathcal{L_{\texttt{KDSP}}} = \sum_{i} \bm{Pr}_{\text{teacher}}(i) \log \frac{\bm{Pr}_{\text{teacher}}(i)}{\bm{Pr}_{\text{student}}(i)},
\end{equation}
where $\bm{Pr}_{\text{teacher}}(i)$ is the probability distribution of the logits of $\bm{V}$ and $\bm{P_s}$, and $\bm{Pr}_{\text{student}}(i)$ is the probability distribution of the logits of $\bm{V}$ and $\bm{T_p}$.
\\
\noindent
The KL divergence term, $\mathcal{L_{\texttt{KDSP}}}$ restricts the model from drifting toward embeddings that are not representative of the actual medical scans. By minimizing this KL divergence, we guide the model to stay within a meaningful embedding space closely related to the content of the medical scans. This alignment helps ensure that the learned embeddings retain essential information about the biomedical images, preventing the model from diverging into unrelated semantic spaces.
\\ \\
The overall learning objective is defined as:
\begin{equation}
    \mathcal{L} = \mathcal{L_{\texttt{CE}}} + \lambda_{1}\mathcal{L_{\texttt{SCCM}}} + \lambda_{2}\mathcal{L_{\texttt{KDSP}}}
\end{equation}
where $\lambda_{1}$, and $\lambda_{2}$ are loss-balancing weights.

%% file: sec/3_results.tex
\section{Experiments and Results}
\label{sec:experiments}
\subsection{Experimental Setup}
We assess the efficacy of our proposed BiomedCoOp framework across a comprehensive set of benchmark biomedical imaging datasets under multiple evaluation protocols designed to test accuracy and generalization within and across various few-shot image classification tasks.
\\
\noindent \textbf{Few-Shot Learning:} To assess the model’s performance under limited supervision, we conduct few-shot experiments with varying numbers of labeled examples per class ($K$ = 1, 2, 4, 8, and 16 shots). This is critical for evaluating the model’s ability to learn effectively from sparse data, a common scenario in biomedical applications, while retaining both task-specific and general domain knowledge.
\\
\noindent \textbf{Base-to-Novel Class Generalization:} To assess model generalizability of our technique, each dataset is divided into base and novel classes. The model is trained on the base classes using a 16-shot setup and subsequently evaluated on both base and novel classes. This setup tests the model’s ability to generalize to unseen classes within the same dataset, showcasing its potential to recognize novel disease presentations without additional fine-tuning.
\\
\noindent \textbf{Datasets:} We conduct our experiments on 11 diverse medical imaging datasets covering 10 different organs and 9 imaging modalities: Computerized Tomography (CTKidney \cite{ctkidney}), Dermatoscopy (DermaMNIST \cite{dermamnist1,dermamnist2}), Endoscopy (Kvasir \cite{kvasir}), Fundus Photography (RETINA \cite{retina1,retina2}), Histopathology (LC25000 \cite{LC25000}, CHMNIST \cite{chmnist}), Magnetic Resonance Imaging (BTMRI \cite{btmri}), Optical Coherence Tomography (OCTMNIST \cite{octmnist}), Ultrasound (BUSI \cite{busi}), and X-Ray (COVID-QU-Ex \cite{covid}, KneeXray \cite{kneexray}). This selection includes complex datasets, such as brain MRI and ultrasound, ensuring a thorough assessment of the model’s performance across a broad spectrum of biomedical imaging contexts. The detailed data split and tasks for the experiments are included in the \textit{Supplementary Materials}.
\input{tables/fewshot_main_results}
\\
\noindent \textbf{Implementation details:} We employed BiomedCLIP with a ViT-B/16 backbone, averaging results over three runs. The training was set to 100 epochs for few-shot and 50 epochs for base-to-novel benchmarking. We initialized the learnable context with the embedding vector corresponding to ``\texttt{a photo of a}" and used 50 LLM prompts, a 0.0025 learning rate, a batch size of 4, and an SGD optimizer across datasets. Optimal values for $\lambda_1$, $\lambda_2$, and $\zeta_s$ were selected using the validation set (detailed values are reported in the \textit{Supplementary Materials}). All experiments ran on a single NVIDIA A100 GPU (40GB RAM).

\subsection{Few-shot Evaluation}
We compared BiomedCoOp against four text-prompt learning methods (CoOp, CoCoOp, ProGrad, KgCoOp), three CLIP-based adapter methods (CLIP-Adapter, Tip-Adapter, Tip-Adapter-F), and two linear probing methods (Standard Linear Probe, LP++). In this study, we focus on text-modal prompt learning techniques that solely optimize text prompts, excluding methods that learn both text and image prompts (i.e. MaPLe, DCPL). Additionally, we limit our comparison to shallow prompt learning techniques rather than deep prompt approaches that require additional parameters for a fair comparison. Additionally, zero-shot and LLM-prompted zero-shot BiomedCLIP configurations were tested. All models used BiomedCLIP as the backbone and were tuned to their optimal settings. As shown in Table~\ref{table:fewshot-main-results}, BiomedCoOp consistently outperformed these baselines, especially in low-shot settings, where it surpassed the second-best performing method- ProGrad- by 5.2\% and 4.6\% in 1- and 2-shot scenarios. This improvement is attributed to BiomedCoOp's use of class-specific, selectively ensembled LLM prompts, enhancing generalizability and sensitivity to diverse biomedical conditions, even with minimal labeled data. BiomedCoOp continued to lead as the number of shots increased, proving its robustness across different data availability scenarios. Its consistent performance across all $K$-shot settings supports BiomedCoOp's effectiveness in prompt-based biomedical adaptation, ensuring reliable cross-dataset accuracy gains.

\subsection{Base-to-Novel Generalization}
We evaluated BiomedCoOp's base-to-novel generalization against the state-of-the-art prompt learning methods (CoOp, CoCoOp, ProGrad, and KgCoOp) by measuring accuracy across both base and novel classes, using harmonic mean (HM) for balanced generalization. \underline{Note that} \textit{the BUSI dataset was excluded due to insufficient class diversity for the experiment}. Results reported in Table~\ref{tab:base-to-new} show that BiomedCoOp consistently outperformed others, with 5-10\% higher accuracy on challenging datasets like CTKidney and Kvasir. This indicates BiomedCoOp’s strong capacity to generalize to unseen biomedical categories—an essential feature in medical contexts with evolving disease patterns. Its LLM-augmented prompts further aid in retaining base knowledge while effectively adapting to novel classes, reducing forgetting.
\input{tables/base2new_main_results}

\subsection{Ablation Experiments}
\subsubsection{Effect of Different Components}
Table \ref{tab:ablation-components} demonstrates the critical contributions from each of BiomedCoOp's components. The baseline BiomedCLIP model, without ensembling or BiomedCoOp components, showed limited accuracy, particularly for novel classes. Adding SCCM alone significantly improved adaptability by embedding contextual medical information, enhancing low-shot performance. KDSP, applied alone, boosted accuracy in novel classes by filtering out low-quality and case-specific prompts. However, using KDSP with a CLIP-only setup hindered generalization, as the model lacked the domain knowledge needed to effectively exclude outliers (gray highlight). The combined use of SCCM and KDSP yielded optimal results (green highlight), balancing generalization and adaptability, especially in few-shot and base-to-novel tasks, affirming both components' essential roles in handling limited-data biomedical scenarios.

\input{tables/component_ablation_study}
\begin{figure}[h]
    \centering
    \includegraphics[width=0.5\textwidth, height=0.35\textwidth]{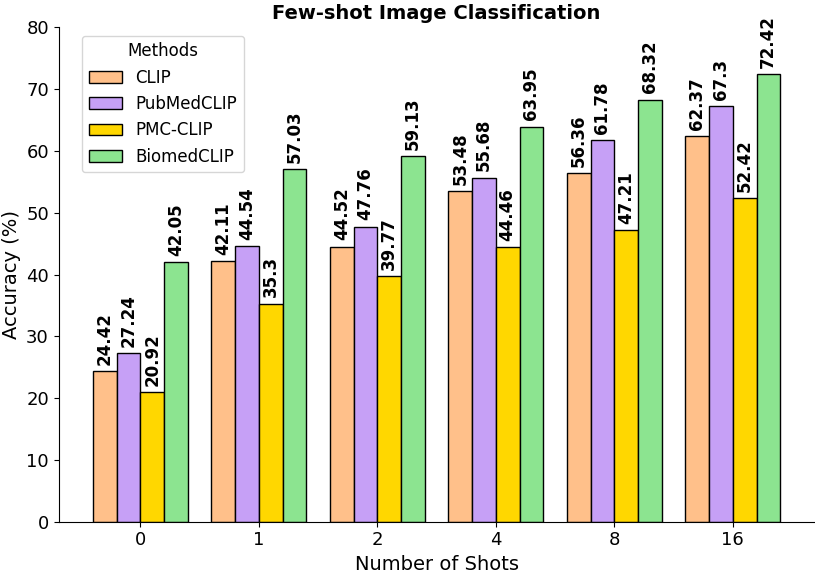}
    \caption{Barplot to compare classification accuracies (\%) of different CLIP-based backbone models in BiomedCoOP across various few-shot settings.}
    \label{fig:model-ablation}
\end{figure}
\input{tables/prompt_ablation}
\begin{figure*}
    \centering
    \includegraphics[height=0.41\textwidth, width=0.75\textwidth]{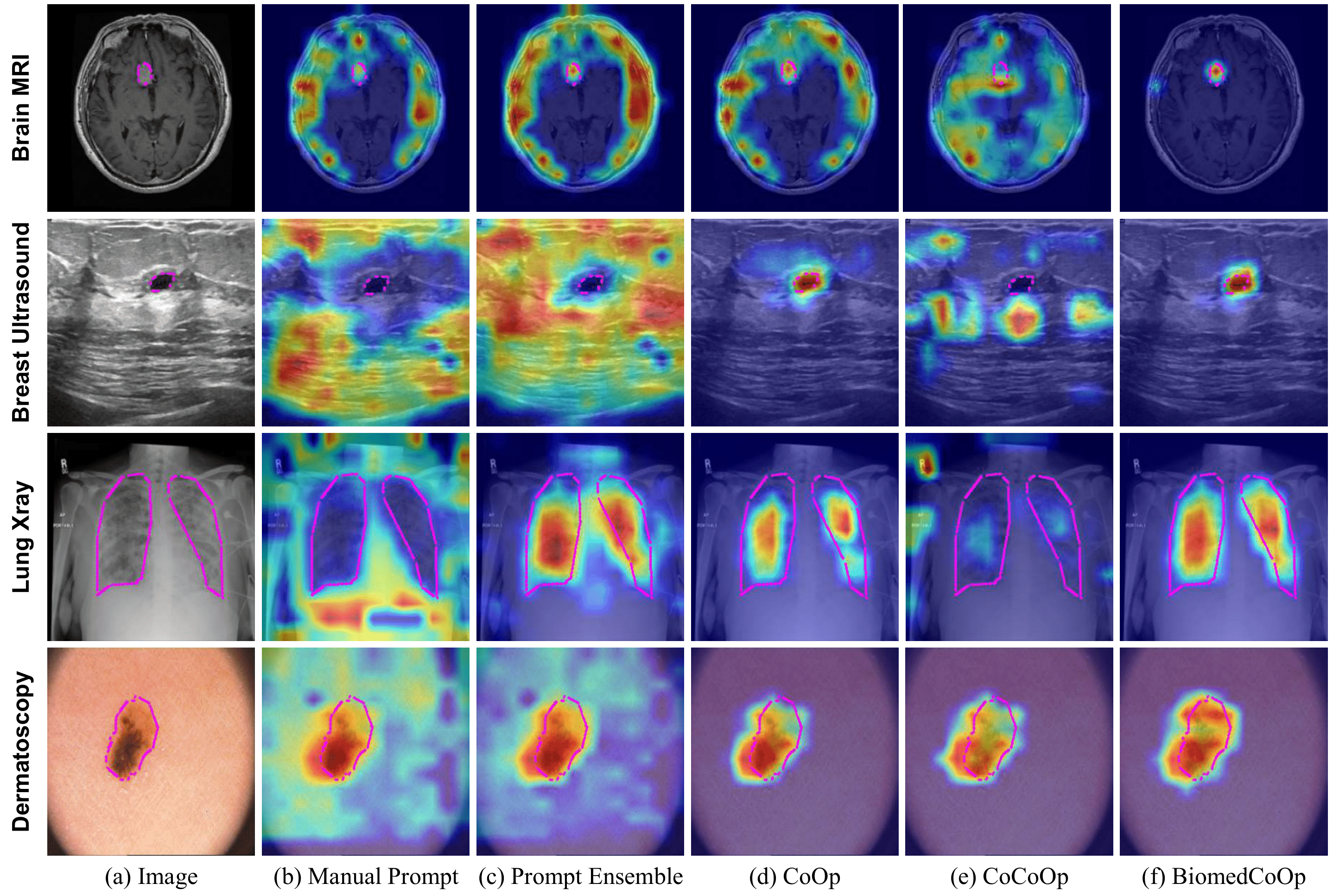}
    \caption{Effect of various text prompt techniques on visual saliency maps. Columns (b)-(f) represent different prompt methods.}
    \label{fig:visual-explainability}
\end{figure*}

\subsubsection{Effect of Number of LLM Prompts}
Prompt diversity in relation to the number of LLM prompts may affect the quality of context mapping. To investigate this, Table \ref{tab:prompt_ablation} shows the effect of increasing LLM-generated prompts on BiomedCoOp’s performance across few-shot settings ($K$ = 0, 1, 2, 4, 8, and 16 shots). At lower shot levels ($K$ = 0 and 1), a higher prompt count noticeably boosts accuracy, improving by 5-6\% as prompts increase from 10 to 50. This indicates that prompt diversity is key for generalization with minimal labeled data. In intermediate shot settings ($K$ = 2 and 4), performance continues to improve with more prompts but at a slower rate, showing diminishing returns as more labeled information becomes available. At higher shot levels (8 and 16), accuracy stabilizes, with only minor gains from additional prompts. As shown in other studies \cite{zaghir2024prompt}, prompt diversity strengthens model performance by providing varied semantic cues that guide the model toward essential biomedical features. Each prompt introduces a unique contextual perspective, enabling the model to build a more robust, flexible understanding of medical concepts. This diversity helps the model focus on shared, critical features, enhancing its ability to recognize subtle variations and generalize effectively, particularly in low-data scenarios.

\subsubsection{Effect of Different CLIP-based Models}
To better understand the impact of backbone CLIP-based models on the proposed BiomedCoOp methods, we examined the performance with four CLIP-based models, including CLIP (ViT-B/16), PubMedCLIP (ViT-B/32), PMC-CLIP (RN50), and BiomedCLIP (ViT-B/16). The bar plot in Fig.~\ref{fig:model-ablation} compares the few-shot classification accuracy of the variants of BiomedCoOp with these CLIP backbones. Notably, BiomedCLIP consistently achieves the highest accuracy across all shot settings, showcasing its ability to capture medical domain-specific features effectively. As the number of shots increases, all models improve, with BiomedCLIP reaching 72.42\% accuracy at 16 shots, significantly outperforming others. CLIP and PMC-CLIP show similar results, while PubMedCLIP remains competitive, but trails BiomedCLIP. These results validate our selection of BiomedCLIP as the VLM backbone and highlight the importance of specialized biomedical VLMs for improving few-shot biomedical image classification performance.

\subsubsection{Visual Interpretability}
In this experiment, gScoreCAM \cite{chen2022gscorecam} was used to assess how different text prompts influence the visual saliency maps of biomedical images. Each column in Fig.~\ref{fig:visual-explainability} represents a distinct prompt type: the ``manual prompt" (column b) uses \texttt{a photo of a [CLASS]}, which tends to make the model focus on global features (i.e. scan modality), rather than the region of interest; the ``prompt ensemble" (column c) averages multiple LLM-generated prompts, potentially introducing conflicting signals and causing the model to focus on the background; and CoOp, CoCoOp, and BiomedCoOp (columns d, e, f) use optimized, learned contextual prompts with the \texttt{[CLASS]} label. Tested on BUSI (breast tumor ultrasound), COVID-QU-Ex (lung X-ray), brain tumor \cite{Cheng2017}, and ISIC \cite{isic1,isic2,isic3} (dermatology) datasets, BiomedCoOp (column f) consistently aligns best with ground truth regions across modalities, accurately highlighting clinically relevant areas, especially in complex modalities like MRI and ultrasound, with notable yet smaller improvements in dermatology due to its similarity to natural images, leading to fewer false positives and negatives. This precise localization enhances interpretability, making it particularly valuable for medical applications \cite{koleilat2024medclip,koleilat2024medclipv2}, where explainability is essential. This observation further validates the effectiveness of our proposed method.

%% file: tables/fewshot_main_results.tex
\begin{table*}[ht]
\centering
\caption{\textbf{Evaluation against state-of-the-art techniques.} This table presents the average classification accuracy (\%) obtained from 11 benchmarks, along with the standard deviation derived from 3 sampled support sets for each dataset. Results show mean$\pm$std. The top-performing results are in bold.}
\tablestyle{-7pt}{1.1}
\addtolength{\tabcolsep}{+20pt}
\resizebox{\textwidth}{!}{%
\begin{tabular}{lcccccc}
\toprule
\textbf{Method} &  $K=1$ & $K=2$  & $K=4$  & $K=8$  & $K=16$ \\
\midrule
\rowcolor{tabhighlight}\multicolumn{6}{c}{Zero-shot Methods} \\
BiomedCLIP \cite{biomedclip} & \multicolumn{5}{c}{$42.05$} \\
BiomedCLIP \cite{biomedclip} + Ensemble & \multicolumn{5}{c}{$52.27$} \\
BiomedCLIP \cite{biomedclip} + Selective Ensemble & \multicolumn{5}{c}{$53.72$} \\
\midrule
\rowcolor{tabhighlight}\multicolumn{6}{c}{CLIP-based Adapter Methods} \\
CLIP-Adapter \cite{gao2024clip} & $44.66$ ± $2.97$ & $43.91$ ± $2.48$ & $44.36$ ± $1.94$ & $45.42$ ± $2.38$ & $46.69$ ± $1.71$ \\
Tip-Adapter \cite{zhang2021tip} & $49.19$ ± $4.84$ & $52.36$ ± $6.57$ & $57.33$ ± $5.07$  & $61.98$ ± $5.76$  & $67.15$ ± $4.25$ \\
Tip-Adapter-F \cite{zhang2021tip} & $51.17$ ± $8.33$ & $52.74$ ± $5.88$ & $61.23$ ± $6.22$ & $65.91$ ± $3.64$ & $70.91$ ± $2.65$ \\
\midrule
\rowcolor{tabhighlight}\multicolumn{6}{c}{Linear Probing Methods} \\
Standard LP \cite{radford2021learning} & $47.25$ ± $8.65$  & $54.21$ ± $7.80$  & $61.00$ ± $6.81$  & $65.85$ ± $4.89$  & $69.40$ ± $2.91$ \\
LP++ \cite{huang2024lp++} & $47.24$ ± $7.68$ & $53.18$ ± $7.29$ & $59.02$ ± $6.93$ & $63.69$ ± $4.68$ & $68.35$ ± $3.59$ \\
\midrule
\rowcolor{tabhighlight}\multicolumn{6}{c}{Prompt Learning Methods} \\
CoOp \cite{zhou2022learning} & $50.16$ ± $6.93$ & $54.18$ ± $4.31$  &  $59.75$ ± $3.72$  & $65.84$ ± $3.66$  &  $69.62$ ± $2.83$ \\
CoCoOp \cite{zhou2022conditional} & $48.49$ ± $4.39$ &  $51.28$ ± $5.06$ &  $54.69$ ± $4.79$  &  $61.08$ ± $3.49$ & $65.09$ ± $2.87$  \\
KgCoOp \cite{yao2023visual} & $50.85$ ± $5.59$ & $53.18$ ± $4.33$ & $57.82$ ± $4.50$ & $62.08$ ± $2.59$ & $62.84$ ± $1.72$ \\
ProGrad \cite{zhu2023prompt} & $51.88$ ± $6.39$ & $54.71$ ± $4.46$ & $60.42$ ± $4.78$ & $65.61$ ± $3.02$ & $67.13$ ± $3.00$ \\
\rowcolor{blue!20} BiomedCoOp (Ours) & \textbf{57.03} ± \textbf{2.80} & \textbf{59.13} ± \textbf{3.64} & \textbf{63.95} ± \textbf{2.42} & \textbf{68.32} ± \textbf{2.65} & \textbf{72.42} ± \textbf{1.69} \\
\bottomrule
\end{tabular}
}
\label{table:fewshot-main-results}
\end{table*}

%% file: tables/base2new_main_results.tex
\begin{table}[h]
\centering
\tablestyle{-18pt}{1.1}
\addtolength{\tabcolsep}{+20pt}
\resizebox{\columnwidth}{!}{%
\begin{tabular}{lc|c c c c c c}
\hline
\multirow{2}{*}{Dataset} &  &  {BiomedCLIP} &  {CoOp} & {CoCoOp} &  {KgCoOp} & 
{ProGrad} & 
{BiomedCoOp} \\
 & &  \cite{biomedclip} & \cite{zhou2022learning} & \cite{zhou2022conditional}  & \cite{kgcoop}  & \cite{prograd}  & (Ours) \\
\midrule
\multirow{3}{*}{\shortstack[l]{Average on\\  10 datasets}}       & Base      & 47.84  & 73.85   & 72.26  & 68.36   & 71.67        & \textbf{76.26} \\
                               & Novel     & 65.42  & 64.75   & 67.03  & 64.08    & 66.93          & \textbf{73.92} \\
                               & HM        & 53.81          & 67.23         & 67.22  & 64.61  & 67.43  & \textbf{75.07} \\
\midrule
\multirow{3}{*}{BTMRI}      & Base   & 40.88          & 82.25         & 77.88  & 78.03  & 82.13          & \textbf{82.42}\\ 
                               & Novel    &  96.18        & 94.51         & 94.84  & 95.05  & 94.98          & \textbf{96.84} \\ 
                               & HM          & 57.37          & 87.95         & 85.53  & 85.69  & 88.09          & \textbf{89.05} \\
\midrule
\multirow{3}{*}{COVID-QU-Ex}    & Base   & 53.96          & 75.92         & \textbf{77.28}  & 75.42     & 75.19  & 75.91\\
                               & Novel     & 89.43          & 90.07         & 87.61   & 89.61  & 90.34          & \textbf{91.63} \\
                               & HM          & 67.31          & 82.39         & 82.12  & 81.90  & 82.07 & \textbf{83.03} \\
\midrule
\multirow{3}{*}{CTKIDNEY}    & Base     & 38.55          & 82.24          & 81.96  & 81.67  & 83.86           & \textbf{86.93} \\
                               & Novel      & 52.99          & 67.92         & 56.56   & 58.45  & 63.01           & \textbf{78.94} \\
                               & HM          & 44.63          & 74.40         & 66.93  & 68.14  & 71.96          & \textbf{82.74} \\
\midrule
\multirow{3}{*}{DermaMNIST}  & Base    & 34.95          & 48.06         & 42.88  & 36.41  & 35.52 & \textbf{54.86}\\
                               & Novel     & 49.59          & 59.41         & 60.66  & 47.31  & 63.28          & \textbf{74.1} \\
                               & HM         & 41.00          & 53.14         & 50.24  & 41.15  & 45.50          & \textbf{63.04} \\
\midrule
\multirow{3}{*}{Kvasir}    & Base     &75.00          & 86.22         & 85.94   & 81.56  & 82.89           & \textbf{86.50}\\
                               & Novel     & 60.50          & 58.06          & 53.95   & 59.00   & 60.45          & \textbf{61.83} \\
                               & HM         & 66.97          & 69.39         & 66.29  & 68.47  & 69.91          & \textbf{72.11} \\
\midrule
\multirow{3}{*}{CHMNIST}       & Base     & 37.63          & \textbf{89.41} & 87.77  & 75.45  & 82.98          & 88.87 \\
                               & Novel     & 40.69          & 35.11         & 42.51  & 38.70   & \textbf{44.19}             & 42.73 \\
                               & HM          & 39.10         & 50.42         & 57.28  & 51.16  & 57.67          & \textbf{57.71} \\
\midrule
\multirow{3}{*}{\shortstack[l]{LC25000}}  & Base    & 59.73          & 90.12         & 88.33  & 88.13  & 90.29          & \textbf{93.77} \\
                               & Novel      & 87.60          & 87.55         & 95.02  & 86.44   & 85.47          & \textbf{97.00} \\
                               & HM         & 71.03          & 88.82         & 91.55  & 87.28  & 87.81          & \textbf{95.36} \\
\midrule
\multirow{3}{*}{RETINA}        & Base    & 45.18          & \textbf{70.98}         & 66.88  & 60.77  & 68.77          & 68.46 \\
                               & Novel      & 55.28 & 56.90         & 65.56  & 54.91     & 58.43          & \textbf{67.72} \\
                               & HM        & 49.72          & 63.16         & 66.21  & 57.69  & 63.18          & \textbf{68.09} \\
\midrule
\multirow{3}{*}{KneeXray}           & Base    & 35.89          & 38.28          & 34.08  & 37.94  & 40.88          & \textbf{44.23} \\
                               & Novel     & 71.90          & 47.69         & 63.14  & 61.19  & 59.12          & \textbf{78.35} \\
                               & HM          & 47.88          & 42.47         & 44.27  & 46.84  & 48.34          & \textbf{56.54} \\
\midrule
\multirow{3}{*}{OCTMNIST}        & Base     & 56.60          & 75.00         & 79.6  & 68.20  & 74.20  & \textbf{80.33}\\
                               & Novel      & 50.00          & 50.23          &\textbf{50.47}     & 50.13   & 50.02          & 50.07 \\
                               & HM          & 53.10          & 60.17         & \textbf{61.77}  & 57.79  & 59.76          & 61.69 \\
\bottomrule
\end{tabular}%
}
    \caption{\small\textnormal{Accuracy comparison (\%) on Base-to-novel generalization of BiomedCoOp with SOTA prompt learning methods}. HM = harmonic mean of the classification accuracy between base and novel classes.}
    \label{tab:base-to-new}
\end{table}

%% file: tables/component_ablation_study.tex
\begin{table}[htpb]
\centering
\newcommand{\cmark}{\ding{51}} 
\newcommand{\xmark}{\ding{55}} 
\tablestyle{-13pt}{1.1}
\arrayrulecolor{black}
\setlength\arrayrulewidth{1pt}
\addtolength{\tabcolsep}{+16pt}
\resizebox{\columnwidth}{!}{%
\begin{tabular}{ccc ccc ccccc }
\toprule
\rowcolor{gray!20} \multicolumn{3}{c}{\textbf{Components}}     & \multicolumn{3}{c}{\textbf{Base-to-Novel}}      & \multicolumn{5}{c}{\textbf{Few-shot}}                                             \\ \midrule
\rowcolor{gray!10} \textbf{BiomedCLIP} & \textbf{SCCM}                   & \textbf{KDSP}                   & \textbf{Base}           & \textbf{Novel}          & \textbf{HM}             & \textbf{1}          & \textbf{2}          & \textbf{4}          & \textbf{8}          & \textbf{16}         \\ \midrule
\xmark & \xmark & \xmark & 71.50          & 45.06          & 55.28          & 41.46          & 45.69          & 51.41          & 54.80          & 61.86          \\
\xmark & \cmark & \xmark & 71.91          & 43.04          & 53.85          & 43.72          & 44.77          & 52.76          & 60.08           & 64.04          \\
\xmark & \xmark & \cmark & 73.21          & 39.95          & 51.69         & 42.14         & 44.27          & 53.75          & 56.67           & 62.14          \\
\rowcolor{gray!20} \xmark & \cmark & \cmark & 72.95          & 39.04          & 50.86          & 42.11          & 44.52          & 53.48          & 56.36           & 62.37          \\
\cmark & \xmark & \xmark & 73.85          & 64.75          & 67.23          & 50.16          & 54.18          & 59.75          & 65.84           & 69.62          \\
\cmark & \cmark & \xmark & 75.21          & 65.79          & 70.19          & 51.62          & 54.99          & 61.43          & 65.93           & 69.99          \\
\cmark & \xmark & \cmark & 75.74          & 72.91          & 74.30          & 56.78          & 58.76          & 63.68          & 67.68           & 71.79          \\
\rowcolor{green!20}\cmark & \cmark & \cmark & \textbf{76.11} & \textbf{73.22} & \textbf{74.64} & \textbf{57.03} & \textbf{59.13} & \textbf{63.95} & \textbf{68.32} & \textbf{72.42} \\ \bottomrule
\end{tabular}%
}
\caption{
Impact of each component of the proposed BiomedCoOp method on the accuracy (\%) of few-shot and Base-to-Novel benchmarks, including BiomedCLIP vs. CLIP, SCCM, and KDSP.
}
\label{tab:ablation-components}
\end{table}

%% file: tables/prompt_ablation.tex
\begin{table}[t!]
   \small \centering
  \setlength{\tabcolsep}{10pt}
    \scalebox{0.75}[0.75]{
    \begin{tabular}{c cccccc}
    \toprule
    Prompts \#  & $K$ = 0 & $K$ = 1 & $K$ = 2 & $K$ = 4 & $K$ = 8 & $K$ = 16\\
            \midrule
    10 & 47.55 & 52.43 & 54.61 & 60.69 & 64.81 & 67.66 \\
    20 & 50.51 & 55.27 & 57.65 & 62.85 & 66.92 & 70.96 \\
    30 & 51.88 & 55.91 & 58.52 & 63.89 & 68.05 & 71.97 \\
    40 & 52.20 & 56.59 & 59.05 & 63.92 & 68.24 & 72.20 \\
    \rowcolor{tabhighlight} 50 & 52.27 & 57.03 & 59.13 & 63.95 & 68.32 & 72.42 \\
    \bottomrule
    \end{tabular}
    }\vspace{-0.5em}
    \caption{BiomedCoOp's classification accuracy (\%) across different K-shot settings with varying numbers of LLM prompts.}
    \label{tab:prompt_ablation}
\end{table}

%% file: sec/4_conclusion.tex
\section{Conclusion}
Our proposed BiomedCoOp framework represents the first large-scale exploration of prompt learning tailored for few-shot adaptation across diverse biomedical datasets. By distilling domain-specific insights from LLMs, BiomedCoOp enriches prompt representations and enhances the model’s ability to generalize across a wide range of biomedical imaging contexts. Our approach combines generalizable biomedical knowledge with selective prompt refinement to improve classification accuracy and generalizability. This study highlights the potential of integrating LLM-derived prompts with BiomedCLIP to achieve accurate, data-efficient biomedical diagnosis, serving as a foundational step towards the broader adoption of adaptable VLMs in clinical applications. 

\noindent\textbf{Acknowledgment} We acknowledge the support of the Natural Sciences and Engineering Research Council of Canada (NSERC).

%% file: sec/X_suppl.tex
\clearpage
\renewcommand{\thetable}{S\arabic{table}}
\renewcommand{\thefigure}{S\arabic{figure}}
\setcounter{table}{0}
\setcounter{figure}{0}

\maketitlesupplementary
\section{Detailed Dataset Overview}
Table \ref{tab:dataset_info} provides a summary of the 11 datasets used for our proposed BiomedCoOp, covering 9 biomedical imaging modalities, such as CT, MRI, X-ray, ultrasound, and others, and 10 different organs. Each dataset is described in terms of its imaging modality, target organ(s), number of classes, and dataset splits (train/validation/test). The datasets span diverse clinical cases, including kidney cysts in CT, various skin lesions in dermatoscopy, and different stages of knee osteoarthritis in X-ray. These datasets capture a wide range of disease classes and imaging types, offering a rich and representative benchmark for biomedical image classification tasks. Instead of using full training splits, we employ random few-shot seeds to ensure efficient and representative learning from limited data. Additionally, the examples for each class are proportionally distributed across the splits, ensuring balanced representation, which enhances model evaluation on clinically relevant data and strengthens BiomedCoOp’s robustness across diverse tasks.

\section{Additional Few-shot Results}
\begin{figure}[h]
    \centering
    \includegraphics[width=0.5\textwidth]{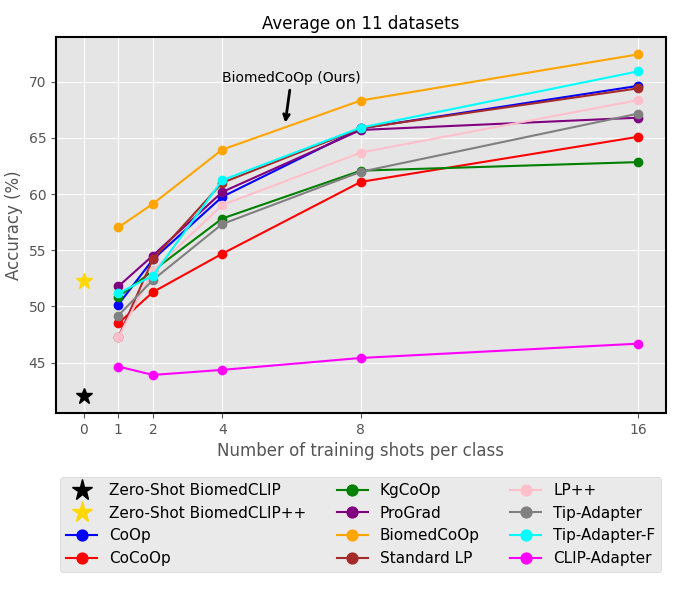}
    \caption{Average classification accuracy (\%) of various few-shot adaptation methods across different numbers of training shots per class.}
    \label{fig:few-shot-main-figure}
\end{figure}

Figure \ref{fig:few-shot-main-figure} demonstrates the performance variations of BiomedCoOp and the baseline models under different few-shot configurations ($K$ = 1,2,4,8,16). It underscores BiomedCoOp’s robustness in adapting to limited data. On the other hand, we provide the detailed few-shot evaluation results for each dataset in Table \ref{tab_appendix:few_shot_experiments}. Overall, BiomedCoOp is on par and regularly outperforms SOTA parameter-efficient techniques across diverse datasets.
\label{sec:fewshot-all-results}
\section{Learnable Context Interpretability}
In this experiment, we aim to investigate the closest words to each of the four learned context tokens in various biomedical datasets, examining how these nearest words align with visual or anatomical characteristics in the images. This could offer some intuitive interpretation for the learned context, which is more abstract than a typical phrase like ``\texttt{A photo of [CLASS]}". As shown in Table \ref{tab:interpret_prompts}, the nearest words to each learned context token are listed, along with their corresponding Euclidean distances in the embedding space (in parentheses). It's particularly intriguing that some learned embeddings capture relevant descriptors, such as ``\texttt{endoscopy}" for Kvasir, ``\texttt{mri}" for BTMRI, or ``\texttt{receptive}" for RETINA, reflecting contextual understanding of these biomedical imaging types.

\section{Effect of Context Length}
As shown in Table \ref{tab:prompt-config}, increasing the context length tends to reduce performance on both base and novel classes. A shorter context length, such as 4, achieves a better balance between base and novel accuracy, resulting in a higher harmonic mean (HM) score. As the context length increases to 16, 32, and 64, the accuracy on novel classes declines more rapidly than on base classes, leading to a sharp reduction in the harmonic mean. This pattern suggests that longer context lengths could diminish the model's ability to generalize effectively across both base and novel classes.
\input{tables/prompt_configurations}
\section{Effect of Prompt Selection Threshold}
\begin{figure}
    \centering
    \includegraphics[width=0.5\textwidth]{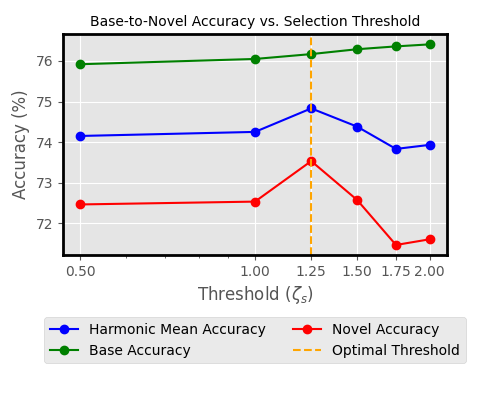}
    \caption{Effect of selection threshold ($\zeta_s$) on Base-to-Novel Generalization}
    \label{fig:threshold-ablation}
\end{figure}

In the \textit{Knowledge Distillation with Selective Prompting} (KDSP) component of our proposed method, we used a statistics-based prompt selection strategy. Figure~\ref{fig:threshold-ablation} illustrates the impact of increasing the selection threshold ($\zeta_s$) for the absolute value of the modified z-score to allow more prompts generated by the LLM to be used. This can lead to an overfitting effect, where the model becomes highly specialized in distinguishing base classes at the expense of its ability to generalize to novel classes. As a result, while the base class accuracy remains high or slightly increases with an increasing threshold, the accuracy for novel classes declines, as shown by the peak in novel accuracy at the optimal threshold ($\zeta_s$=1.25). Beyond this point, higher thresholds reduce generalization capability, leading to decreased harmonic mean accuracy.

\section{Selective Prompting for SCCM}
We didn't perform prompt selection in the \textit{Semantic Consistency by Contextual Mapping} (SCCM) component of our proposed BiomedCoOp framework. To verify the effect of prompt selection for SCCM, we compare the model performance with and without prompt selection. Table \ref{tab:sccm-selective} shows that excluding outliers during contextual mapping marginally improves accuracy on base classes (76.26\% to 76.39\%), but reduces accuracy on novel classes (73.92\% to 72.59\%), resulting in a lower harmonic mean (74.44\% to 73.92\%).  This suggests that outlier exclusion when applied to the SCCM component causes the prompts to overfit to base classes, reducing their flexibility and hindering generalization to novel classes. Thus, keeping all prompt samples in the mapping process (i.e., SCCM) helps maintain broader generalization, balancing performance across both base and novel classes.
\section{Additional Comparisons with Other Recent Methods}
We compared our method with two more recent SOTA CLIP adaptation methods (XCoOp \cite{bie2024xcoop} and DCPL \cite{cao2024domain}) on all datasets in Tables \ref{table:fewshot-additional} and \ref{tab:base2new-additional}. We also compare with zero-shot methods (in blue). All methods were tuned to their optimal settings for each dataset, and 16-shot setting is used for all in Table \ref{tab:base2new-additional}. Specifically, we used an alternate version of DCPL, denoted DCPL*, which uses BiomedCLIP as the LSDM with deep multimodal prompting, whereas XCoOp directly utilizes the BiomedCLIP backbone. Our results demonstrate that our method consistently outperforms these approaches. On the other hand, average ensemble of LLM prompts improves zero-shot classification, but is suboptimal to prompt selection (Table \ref{table:fewshot-additional}).
\input{tables/fewshot_additional}
\input{tables/base2new_additional}

\section{Effect of LLM used}
Table \ref{table:llm-ablation} presents an ablation study using 50 prompts from three recent LLMs across three datasets under the 4-shot evaluation. The results show that our model is robust to different text distributions, even with smaller LLMs like Gemma-2-2b, highlighting the benefit of our selection strategy in learning vigorous representations.
\input{tables/llm_ablation}

\input{tables/selective_sccm}
\section{Additional Hyperparameters}
Table \ref{tab:hyperparameters} outlines the selected hyperparameters ($\lambda_1$, $\lambda_2$, and $\zeta_s$) used across various datasets for BiomedCoOp’s few-shot and base-to-novel benchmarks. These parameters were optimized to balance classification accuracy and model adaptability, with $\lambda_1$ and $\lambda_2$ controlling the weight of the consistency and distillation losses, respectively, and $\zeta_s$ setting the selection threshold for prompt refinement. The selection threshold $\zeta_s$ remains consistently in the range [1.25, 1.5] across most datasets, indicating a stable value for effective prompt selection.
\input{tables/hyperparameters}

\section{LLM Prompts Used}
We include here one text prompt generated from GPT-4 for each class across all the datasets:
\\ \\
{
``\texttt{The image of a normal brain on MRI shows a clear differentiation between different brain regions with no disruptions.}"\\ \\
\noindent
``\texttt{Central necrosis and surrounding edema in glioma tumor on MRI scan.}"\\ \\
\noindent
``\texttt{Meningioma tumor on MRI displaying a dural tail sign and homogeneous enhancement.}"\\ \\
\noindent
``\texttt{Pituitary tumors often cause sellar expansion and may invade adjacent structures.}"\\ \\
\noindent
``\texttt{A routine ultrasound showing a hypoechoic, well-defined nodule, indicating a benign breast tumor.}"\\ \\
\noindent
``\texttt{An ultrasound revealing microcalcifications within the mass, indicating a malignant breast tumor.}"\\ \\
\noindent
``\texttt{A grayscale ultrasound highlighting well-defined ducts and lobules, characteristic of a normal breast ultrasound scan.}"\\ \\
\noindent
``\texttt{An X-ray scan showing bilateral airspace consolidation, typical of covid lungs.}"\\ \\
\noindent
``\texttt{A chest X-ray image with reticular and nodular opacities, indicative of lung opacity lungs.}"\\ \\
\noindent
``\texttt{An X-ray revealing no signs of consolidation or effusion, suggesting normal lungs.}"\\ \\
\noindent
``\texttt{An X-ray image revealing multifocal ground-glass and consolidative opacities, indicative of viral pneumonia lungs.}"\\ \\
\noindent
``\texttt{A CT image showing a lesion with uniform density and no internal irregularities, indicative of a cyst kidney.}"\\ \\
\noindent
``\texttt{A CT scan showing a calcified structure with acoustic shadowing, consistent with a kidney stone.}"\\ \\
\noindent
``\texttt{A CT scan showing a lesion with poorly defined margins, consistent with a kidney tumor.}"\\ \\
\noindent
``\texttt{A CT image revealing no signs of renal atrophy or cortical thinning, suggesting a normal kidney.}"\\ \\
\noindent
``\texttt{Actinic keratosis lesions may become thicker and more pronounced over time without treatment.}"\\ \\
\noindent
``\texttt{BCC lesions may bleed with minor trauma, such as shaving, due to their friable nature.}"\\ \\
\noindent
``\texttt{Cryotherapy, using liquid nitrogen, is a common treatment for seborrheic keratosis, causing the lesions to blister and fall off.}"\\ \\
\noindent
``\texttt{Dermatofibromas can be multiple in patients with systemic lupus erythematosus or other autoimmune conditions.}"\\ \\
\noindent
``\texttt{A clinical image with a lesion that has changed in size or texture, indicative of melanoma.}"\\ \\
\noindent
``\texttt{Melanocytic nevi can become darker and larger during pregnancy due to hormonal changes and increased melanin production.}"\\ \\
\noindent
``\texttt{The diagnosis of vascular lesions often requires a combination of clinical examination and sometimes imaging studies.}"\\ \\
\noindent
``\texttt{Dyed lifted polyps can exhibit various morphological features, including lobulated, sessile, or pedunculated appearances.}"\\ \\
\noindent
``\texttt{Endoscopic images of dyed resection margins often show a bright, distinct color outlining the area of resection, contrasting with the surrounding mucosa.}"\\ \\
\noindent
``\texttt{In severe cases, esophagitis may lead to strictures or narrowing of the esophageal lumen, visible during endoscopy.}"\\ \\
\noindent
``\texttt{Endoscopic images of the normal cecum show a well-defined junction with the ascending colon, without any transitional abnormalities.}"\\ \\
\noindent
``\texttt{Endoscopic examination of the normal pylorus shows a lack of any masses, polyps, or other abnormal growths.}"\\ \\
\noindent
``\texttt{The Z line in a normal endoscopy appears intact and well-defined, with no evidence of structural compromise.}"\\ \\
\noindent
``\texttt{Polyps can be classified based on their appearance and histological features, including adenomatous polyps, hyperplastic polyps, or inflammatory polyps.}"\\ \\
\noindent
``\texttt{Ulcerative colitis can be associated with extra-intestinal manifestations, including dermatological, joint, ocular, or hepatobiliary complications.}"\\ \\
}
\input{tables/datasets.tex}
\input{tables/fewshot_all_results}
\input{tables/interpret_prompts}

\clearpage

%% file: tables/prompt_configurations.tex
\begin{table}[htpb]
\centering
\newcommand{\cmark}{\ding{51}} 
\newcommand{\xmark}{\ding{55}} 
\tablestyle{-10pt}{1.1}
\addtolength{\tabcolsep}{+18pt}
\resizebox{\columnwidth}{!}{%
\begin{tabular}{c|ccc}
\toprule
 \textbf{Context Length}                   & \textbf{Base Acc.}           & \textbf{Novel Acc.}          & \textbf{HM}    \\ \midrule
\rowcolor{tabhighlight} 4 & 76.11         & 73.22          &   74.64        \\
16 & 74.93          & 67.98          &  71.29         \\
32 &  72.34         &  62.73         &     67.19      \\
64 &    71.50       &    58.99       &     64.65      \\
\bottomrule
\end{tabular}%
}
\caption{
Effect of the context vector length on classification accuracy (\%) in Base-to-Novel generalization.
}
\label{tab:prompt-config}
\end{table}

%% file: tables/fewshot_additional.tex
\begin{table}[ht]
\centering
\caption{Avg. accuracy (\%) comparison of additional SOTA methods in few-shot learning. DCPL* utilizes BiomedCLIP as the LSDM for domain knowledge while XCoOp directly utilizes the BiomedCLIP backbone. Methods in blue use zero-shot setting.}
\tablestyle{-7pt}{1.1}
\addtolength{\tabcolsep}{+10pt}
\resizebox{0.5\textwidth}{!}{%
\begin{tabular}{lccccc} 
\toprule
\textbf{Method} &  $K=1$ & $K=2$  & $K=4$  & $K=8$  & $K=16$ \\
\midrule
\textcolor{blue}{BiomedCLIP} & \multicolumn{5}{c}{$42.05$} \\
\textcolor{blue}{BiomedCLIP + Ensemble} & \multicolumn{5}{c}{$52.27$ $(+10.22$ from 
 BiomedCLIP)} \\
\textcolor{blue}{BiomedCLIP + Selection} & \multicolumn{5}{c}{$53.72$ $(+11.67$ from 
 BiomedCLIP)} \\
DCPL*  & $45.65_{8.86}$ & $51.65_{8.79}$ & $56.62_{7.51}$  & $62.85_{8.40}$   & $68.79_{4.80}$   \\
XCoOp & $52.50_{5.91}$ &  $55.39_{5.74}$ &  $60.87_{4.18}$  & $66.37_{3.44}$  &  $71.04_{1.95}$ \\
BiomedCoOp (Ours) & \textbf{$57.03$}$_{\textbf{$2.80$}}$ & \textbf{$59.13$}$_{\textbf{$3.64$}}$ & \textbf{$63.95$}$_{\textbf{$2.42$}}$ & \textbf{$68.32$}$_{\textbf{$2.65$}}$ & \textbf{$72.42$}$_{\textbf{$1.69$}}$ \\
\bottomrule
\end{tabular}
}
\label{table:fewshot-additional}
\end{table}

%% file: tables/base2new_additional.tex
\begin{table}[htpb]
\centering
\caption{Base-to-novel generalization for recent SOTA methods. DCPL* utilizes BiomedCLIP as the LSDM for domain knowledge while XCoOp directly utilizes the BiomedCLIP backbone.}
\newcommand{\cmark}{\ding{51}} 
\newcommand{\xmark}{\ding{55}} 
\begin{tabular}{c|ccc} 
\toprule
 \textbf{Method (K=16)} & \textbf{Base Acc.} & \textbf{Novel Acc.} & \textbf{HM} \\ 
\midrule
XCoOp &  74.62\%     &   63.19\%      &   68.43\%       \\
DCPL* & 68.70\% & 40.35\% & 50.84\% \\
BiomedCoOp (Ours)  & \textbf{76.26\%} & \textbf{73.92\%} & \textbf{75.07\%} \\
\bottomrule
\end{tabular}%
\label{tab:base2new-additional}
\end{table}

%% file: tables/llm_ablation.tex
\begin{table}[ht]
\centering
\caption{Effect of different LLM choices on 4-shot accuracy (\%)}
\tablestyle{-7pt}{1.0} 
\setlength{\tabcolsep}{3pt} 
\scriptsize
\resizebox{0.48\textwidth}{!}{%
\begin{tabular}{lccc}
\toprule
\textbf{Method} &  BTMRI & COVID-QU-Ex  & CTKidney \\
\midrule
LLaMA-3-8b & 76.61$_{3.53}$ &  73.20$_{1.84}$ &  \textbf{67.81}$_{\textbf{0.39}}$ \\
Gemma-2-2b  & 76.69$_{3.46}$ &  72.46$_{2.46}$ &  66.03$_{0.96}$ \\
GPT-4  & \textbf{77.23}$_{\textbf{3.90}}$ &  \textbf{73.28}$_{\textbf{2.30}}$ &  66.50$_{1.92}$ \\
\bottomrule
\end{tabular}
}
\label{table:llm-ablation}
\end{table}

%% file: tables/selective_sccm.tex
\begin{table}[htpb]
\centering
\newcommand{\cmark}{\ding{51}} 
\newcommand{\xmark}{\ding{55}} 
\tablestyle{-10pt}{1.1}
\addtolength{\tabcolsep}{+15pt}
\resizebox{\columnwidth}{!}{%
\begin{tabular}{c|ccc}
\toprule
 \textbf{Component}                   & \textbf{Base Acc.}           & \textbf{Novel Acc.}          & \textbf{HM}    \\ \midrule
\rowcolor{tabhighlight} SCCM without SPOE & 76.26         & 73.92          &   75.07        \\
SCCM with SPOE & 76.39          & 72.59          &  74.44         \\
\bottomrule
\end{tabular}%
}
\caption{
Effect of excluding outliers in the SCCM block on classification accuracy (\%) in Base-to-Novel generalization. SPOE = Selective Prompting via Outlier Exclusion. HM = harmonic mean.
}
\label{tab:sccm-selective}
\end{table}

%% file: tables/hyperparameters.tex
\begin{table}[h]
\centering
\tablestyle{-12pt}{1.1}
\addtolength{\tabcolsep}{+20pt}
\resizebox{\columnwidth}{!}{%
\begin{tabular}{lc|c c c}
\hline
{Dataset} & Benchmark & {$\lambda_1$} &  {$\lambda_2$} & {$\zeta_s$}  \\
\midrule
\multirow{2}{*}{BTMRI}     & Few-shot & 0.5  & 0.25   & 1.5  \\
                           &  Base-to-Novel   & 0.5  & 0.5   & 1.25  \\
\midrule
\multirow{2}{*}{BUSI}     & Few-shot & 0.75  & 0.75   & 1.5  \\
                           &  Base-to-Novel   & \textbf{-}  & \textbf{-}   & \textbf{-}  \\
\midrule
\multirow{2}{*}{COVID-QU-Ex}     & Few-shot & 0.5  & 2.0   & 1.5  \\
                           &  Base-to-Novel  & 20.0  & 1.0  & 1.25 \\
\midrule
\multirow{2}{*}{CTKIDNEY}     & Few-shot & 1.0  & 0.5   & 1.5  \\
                           &  Base-to-Novel   & 10.0  & 0.25   & 1.25  \\
\midrule
\multirow{2}{*}{DermaMNIST}     & Few-shot & 5.0  & 20.0   & 1.5  \\
                           &  Base-to-Novel   & 2.0  & 0.5   & 1.5  \\
\midrule
\multirow{2}{*}{Kvasir}     & Few-shot & 0.75  & 0.75   & 1.5  \\
                           &  Base-to-Novel   & 1.0  & 1.0   & 1.25  \\
\midrule
\multirow{2}{*}{CHMNIST}     & Few-shot & 0.25  & 0.25   & 1.5  \\
                           &  Base-to-Novel   & 10.0  & 1.0   & 1.5  \\
\midrule
\multirow{2}{*}{LC25000}     & Few-shot & 0.5  & 0.5   & 1.5  \\
                           &  Base-to-Novel   & 0.25  & 0.75   & 1.25  \\
\midrule
\multirow{2}{*}{RETINA}     & Few-shot & 0.25  & 0.25   & 1.5  \\
                           &  Base-to-Novel   & 5.0  & 1.0   & 2.0  \\
\midrule
\multirow{2}{*}{KneeXray}     & Few-shot & 5.0  & 20.0   & 1.75  \\
                           &  Base-to-Novel   & 0.25  & 3.0   & 1.25  \\
\midrule
\multirow{2}{*}{OCTMNIST}     & Few-shot & 1.0  & 0.75   & 1.5  \\
                           &  Base-to-Novel   & 0.75  & 0.5   & 1.5  \\
\bottomrule
\end{tabular}%
}
    \caption{\small\textnormal{Hyperparameter values for $\lambda_1$, $\lambda_2$, and $\zeta_s$ across different datasets and benchmarks.}}
    \label{tab:hyperparameters}
\end{table}

%% file: tables/datasets.tex
\begin{table*}[h]
\tablestyle{-27pt}{1.1}
\addtolength{\tabcolsep}{+30pt}
\resizebox{\textwidth}{!}{%
\begin{tabular}{|c|c|c|c|c|}
\hline
\textbf{Modality}                            & \textbf{Organ(s)}                                    & \textbf{Name} & \textbf{Classes}                                                                                                                                                                                  & \textbf{\# train/val/test} \\ \hline
Computerized Tomography                          & Kidney                                               & CTKidney \cite{ctkidney}     & \begin{tabular}[c]{@{}c@{}} Kidney Cyst, Kidney Stone, \\ Kidney Tumor, Normal Kidney\end{tabular} & 6221/2487/3738             \\ \hline
Dermatoscopy                                   & Skin                                                 & DermaMNIST \cite{dermamnist1,dermamnist2}   & \begin{tabular}[c]{@{}c@{}}Actinic Keratosis, Basal Cell Carcinoma, \\ Benign Keratosis, Dermatofibroma, \\ Melanocytic nevus, Melanoma, Vascular Lesion\end{tabular} & 7007/1003/2005 \\ \hline
Endoscopy                                    & Colon                                                & Kvasir \cite{kvasir}        & \begin{tabular}[c]{@{}c@{}}Dyed Lifted Polyps, Normal Cecum, \\ Esophagitis, Dyed Resection Margins, \\ Normal Pylorus, Normal Z Line, \\ Polyps, Ulcerative Colitis\end{tabular}                 & 2000/800/1200              \\ \hline
Fundus Photography               & Retina                                                  & RETINA \cite{retina1,retina2}       & \begin{tabular}[c]{@{}c@{}}Cataract, Diabetic Retinopathy, \\ Glaucoma, Normal Retina\end{tabular}                                                                                                & 2108/841/1268              \\ \hline
\multirow{5}{*}{Histopathology}                   & \begin{tabular}[c]{@{}c@{}}Lung\\ Colon\end{tabular} & LC25000 \cite{LC25000}      & \begin{tabular}[c]{@{}c@{}}Colon Adenocarcinoma, Colon Benign Tissue, \\ Lung Adenocarcinoma, Lung Benign Tissue, \\ Lung Squamous Cell Carcinoma\end{tabular}                                    & 12500/5000//7500           \\ \cline{2-5} 
                                             & Colorectal                                           & CHMNIST \cite{chmnist}      & \begin{tabular}[c]{@{}c@{}}Adipose Tissue, Complex Stroma, \\ Debris, Empty Background, \\ Immune Cells, Normal Mucosal Glands, \\ Simple Stroma, Tumor Epithelium\end{tabular}                   & 2496/1000/1504             \\ \hline
\multirow{1}{*}{Magnetic Resonance Imaging} & \multirow{1}{*}{Brain}                               & BTMRI \cite{btmri}         & \begin{tabular}[c]{@{}c@{}}Glioma Tumor, Meningioma Tumor, \\ Normal Brain, Pituitary Tumor\end{tabular}                                                                                          & 2854/1141/1717            \\ \hline
Optical Coherence Tomography               & Retina                                                  & OCTMNIST \cite{octmnist}       & \begin{tabular}[c]{@{}c@{}}Choroidal Neovascularization, Drusen, \\ Diabetic Macular Edema, Normal \end{tabular}                                                                                                & 97477/10832/1000   \\ \hline
Ultrasound                                   & Breast                                               & BUSI \cite{busi}         & \begin{tabular}[c]{@{}c@{}}Benign Tumors, Malignant Tumors, \\ Normal Scans\end{tabular}                                                                                                          & 389/155/236                \\ \hline
\multirow{3}{*}{X-Ray}                       & Chest                                                & COVID-QU-Ex \cite{covid}  & \begin{tabular}[c]{@{}c@{}}COVID-19, Lung Opacity, \\ Normal Lungs, Viral Pneumonia\end{tabular}                                                                                                  & 10582/4232/6351           \\ \cline{2-5} 
                                             & Knee                                                 & KneeXray \cite{kneexray}     & \begin{tabular}[c]{@{}c@{}}No, Doubtful, Minimal, \\ Moderate, and Severe Osteoarthritis\end{tabular}                                                                                             & 5778/826/1656              \\ \hline
\end{tabular}
}
\caption{An overview of the 11 datasets used spanning 9 biomedical imaging modalities and 10 different organs.}%
\label{tab:dataset_info}
\end{table*}
        

%% file: tables/fewshot_all_results.tex
\begin{table*}[ht]
\centering
\tablestyle{-5pt}{1.1}
\addtolength{\tabcolsep}{+20pt}
\resizebox{\textwidth}{!}{%
\begin{tabular}{ll|ccccc}
\toprule
\textbf{Dataset} & 
 \textbf{Method} & 
$K$ = 1 &
$K$ = 2 & 
$K$ = 4 & 
$K$ = 8 &
$K$ = 16 \\  \midrule
\multirow{12}{*}{BTMRI}      & BiomedCLIP  & \multicolumn{5}{c}
{$56.79$}\\ 
& BiomedCLIP + Ensemble & \multicolumn{5}{c}{$61.04$} \\
& CLIP-Adapter  & $56.80_{\pm 0.48}$ & $57.13_{\pm 0.88}$ & $56.80_{\pm 0.48}$ & $57.15_{\pm 0.91}$ & $60.16_{\pm 0.32}$ \\
& Tip-Adapter  & $66.66_{\pm 4.37}$ & $67.77_{\pm 2.74}$ & $76.37_{\pm 1.69}$  & $73.75_{\pm 3.15}$  & $78.97_{\pm 1.25}$ \\
& Tip-Adapter-F  & $59.60_{\pm 2.28}$ & $61.94_{\pm 6.74}$ & $77.90_{\pm 1.71}$ & $79.18_{\pm 1.80}$ & $82.27_{\pm 2.33}$ \\
& Standard LP  & $62.24_{\pm 5.03}$  & $72.45_{\pm 5.27}$  & $75.98_{\pm 1.94}$  & $77.63_{\pm 3.45}$  & $81.24_{\pm 2.56}$ \\
& LP++  & $64.72_{\pm 6.16}$ & $71.69_{\pm 5.88}$ & $75.48_{\pm 1.41}$ & $77.11_{\pm 1.28}$ & $81.61_{\pm 1.31}$ \\
& CoOp  & $63.82_{\pm 3.94}$ &  $68.82_{\pm 5.15}$ &  $74.68_{\pm 2.99}$  &  $79.27_{\pm 1.9}$ & $82.37_{\pm 1.89}$  \\
& CoCoOp  & $59.47_{\pm 0.78}$ &  $64.14_{\pm 0.64}$ &  $67.83_{\pm 4.8}$  &  $71.69_{\pm 4.4}$ & $78.45_{\pm 1.83}$  \\
& KgCoOp  & $63.33_{\pm 3.66}$ & $70.16_{\pm 5.47}$ & $75.4_{\pm 2.45}$ & $79.79_{\pm 0.99}$ & $81.07_{\pm 0.33}$ \\
& ProGrad  & $66.92_{\pm 2.10}$ & $71.46_{\pm 3.46}$ & $76.24_{\pm 5.07}$ & $78.82_{\pm 1.77}$ & $82.84_{\pm 1.02}$ \\
\rowcolor{tabhighlight} &  BiomedCoOp (Ours) & $65.08_{\pm 1.81}$ & $70.57_{\pm 4.31}$ & $77.23_{\pm 3.9}$ & $78.55_{\pm 2.19}$ & $83.3_{\pm 1.34}$ \\
\midrule
\multirow{12}{*}{BUSI}      & BiomedCLIP  & \multicolumn{5}{c}{$59.75$}\\ 
& BiomedCLIP + Ensemble & \multicolumn{5}{c}{$59.75$} \\
& CLIP-Adapter  & $61.44_{\pm 0.78}$ & $61.01_{\pm 1.03}$ & $61.72_{\pm 0.81}$ & $61.86_{\pm 1.41}$ & $63.55_{\pm 2.17}$ \\
& Tip-Adapter  & $62.71_{\pm 2.56}$ & $61.44_{\pm 2.44}$ & $59.03_{\pm 1.13}$  & $55.93_{\pm 11.37}$  & $68.78_{\pm 5.54}$ \\
& Tip-Adapter-F  & $61.86_{\pm 2.17}$ & $56.35_{\pm 7.25}$ & $64.54_{\pm 7.01}$ & $68.50_{\pm 2.26}$ & $71.89_{\pm 1.25}$ \\
& Standard LP  & $51.41_{\pm 10.78}$  & $47.88_{\pm 6.44}$  & $53.38_{\pm 7.12}$  & $65.53_{\pm 6.34}$  & $68.78_{\pm 1.80}$ \\
& LP++  & $51.12_{\pm 4.95}$ & $55.50_{\pm 2.38}$ & $60.31_{\pm 3.42}$ & $66.10_{\pm 2.34}$ & $70.05_{\pm 1.58}$ \\
& CoOp  & $48.73_{\pm 3.3}$ & $53.53_{\pm 2.8}$ & $60.17_{\pm 3.65}$ & $64.69_{\pm 6.4}$ & $69.49_{\pm 3.3}$ \\
& CoCoOp  & $52.26_{\pm 3.73}$ &  $49.15_{\pm 2.77}$ &  $59.75_{\pm 1.83}$  &  $65.82_{\pm 3.83}$ & $70.2_{\pm 1.22}$  \\
& KgCoOp  & $53.39_{\pm 7.25}$ & $55.51_{\pm 3.30}$ & $62.01_{\pm 4.38}$ & $67.37_{\pm 2.42}$ & $70.62_{\pm 2.11}$ \\
& ProGrad  & $46.33_{\pm 4.23}$ & $49.15_{\pm 7.32}$ & $62.29_{\pm 7.49}$ & $64.83_{\pm 4.20}$ & $71.47_{\pm 2.69}$ \\
\rowcolor{tabhighlight} &  BiomedCoOp (Ours) & $50.71_{\pm 1.74}$ & $50.71_{\pm 7.34}$ & $59.32_{\pm 1.04}$ & $63.27_{\pm 4.61}$ & $70.34_{\pm 2.27}$ \\
\midrule
\multirow{12}{*}{COVID-QU-Ex}      & BiomedCLIP  & \multicolumn{5}{c}{$43.8$}\\ 
& BiomedCLIP + Ensemble & \multicolumn{5}{c}{$66.86$} \\
& CLIP-Adapter  & $50.42_{\pm 1.55}$ & $43.04_{\pm 1.16}$ & $46.28_{\pm 3.30}$ & $48.68_{\pm 1.13}$ & $49.55_{\pm 1.35}$ \\
& Tip-Adapter  & $62.13_{\pm 7.82}$ & $58.72_{\pm 5.19}$ & $63.84_{\pm 10.41}$  & $66.77_{\pm 5.64}$  & $73.05_{\pm 1.04}$ \\
& Tip-Adapter-F  & $54.89_{\pm 17.51}$ & $54.01_{\pm 7.87}$ & $69.97_{\pm 4.13}$ & $69.89_{\pm 4.08}$ & $76.07_{\pm 3.22}$ \\
& Standard LP  & $49.91_{\pm 10.98}$  & $48.06_{\pm 16.94}$  & $60.55_{\pm 13.60}$  & $68.29_{\pm 6.12}$  & $71.98_{\pm 1.88}$ \\
& LP++  & $46.41_{\pm 10.75}$ & $56.42_{\pm 15.04}$ & $62.32_{\pm 9.54}$ & $66.19_{\pm 8.40}$ & $72.79_{\pm 1.17}$ \\
& CoOp  & $58.82_{\pm 14.51}$ & $58.37_{\pm 8.14}$  &  $67.03_{\pm 6.58}$  & $74.66_{\pm 0.29}$  &  $76.37_{\pm 1.39}$ \\
& CoCoOp  & $69.36_{\pm 2.79}$ &  $68.8_{\pm 2.65}$ &  $63.7_{\pm 10.27}$  &  $69.36_{\pm 3.28}$ & $74.52_{\pm 0.72}$  \\
& KgCoOp  & $61.68_{\pm 9.84}$ & $54.68_{\pm 12.19}$ & $65.91_{\pm 8.61}$ & $74.86_{\pm 0.28}$ & $75.65_{\pm 0.88}$ \\
& ProGrad  & $60.42_{\pm 11.74}$ & $64.22_{\pm 6.44}$ & $68.56_{\pm 3.2}$ & $74.65_{\pm 1.09}$ & $74.93_{\pm 1.07}$ \\
\rowcolor{tabhighlight} &  BiomedCoOp (Ours) & $72.64_{\pm 2.41}$ & $71.53_{\pm 1.5}$ & $73.28_{\pm 2.30}$ & $76.26_{\pm 0.38}$ & $78.72_{\pm 0.23}$ \\
\midrule
\multirow{12}{*}{CTKIDNEY}      & BiomedCLIP  & \multicolumn{5}{c}{$42.43$}\\ 
& BiomedCLIP + Ensemble & \multicolumn{5}{c}{$56.82$} \\
& CLIP-Adapter  & $47.17_{\pm 3.74}$ & $41.94_{\pm 2.15}$ & $42.19_{\pm 2.27}$ & $44.64_{\pm 0.90}$ & $47.28_{\pm 1.41}$ \\
& Tip-Adapter  & $45.85_{\pm 5.41}$ & $51.65_{\pm 7.87}$ & $55.33_{\pm 4.10}$  & $69.89_{\pm 8.74}$  & $73.38_{\pm 7.77}$ \\
& Tip-Adapter-F  & $46.68_{\pm 6.70}$ & $58.99_{\pm 8.54}$ & $60.18_{\pm 10.73}$ & $75.24_{\pm 6.89}$ & $82.07_{\pm 3.29}$ \\
& Standard LP  & $43.82_{\pm 6.43}$  & $59.35_{\pm 6.49}$  & $69.54_{\pm 7.67}$  & $78.89_{\pm 7.37}$  & $82.50_{\pm 5.22}$ \\
& LP++  & $57.70_{\pm 2.85}$ & $61.57_{\pm 3.38}$ & $65.73_{\pm 9.15}$ & $77.06_{\pm 7.96}$ & $79.07_{\pm 7.67}$ \\
& CoOp  & $54.51_{\pm 8.74}$ & $60.57_{\pm 2.26}$  &  $68.12_{\pm 2.11}$  & $77.4_{\pm 3.87}$  &  $83.52_{\pm 1.8}$ \\
& CoCoOp  & $47.88_{\pm 7.72}$ &  $52.71_{\pm 9.71}$ &  $61.07_{\pm 1.33}$  &  $73.93_{\pm 1.5}$ & $77.7_{\pm 2.65}$  \\
& KgCoOp  & $58.92_{\pm 1.28}$ & $62.81_{\pm 3.38}$ & $68.68_{\pm 5.54}$ & $77.43_{\pm 4.2}$ & $77.67_{\pm 3.12}$ \\
& ProGrad  & $54.65_{\pm 8.97}$ & $64.66_{\pm 5.31}$ & $67.90_{\pm 2.02}$ & $78.23_{\pm 4.74}$ & $81.13_{\pm 2.28}$ \\
\rowcolor{tabhighlight} &  BiomedCoOp (Ours) & $56.13_{\pm 4.19}$ & $64.21_{\pm 5.57}$ & $66.5_{\pm 1.92}$ & $77.16_{\pm 3.98}$ & $83.20_{\pm 2.37}$ \\
\midrule
\multirow{12}{*}{DermaMNIST}      & BiomedCLIP  & \multicolumn{5}{c}{$38.75$}\\ 
& BiomedCLIP + Ensemble & \multicolumn{5}{c}{$53.62$} \\
& CLIP-Adapter  & $35.96_{\pm 6.70}$ & $36.01_{\pm 6.63}$ & $34.97_{\pm 4.17}$ & $34.28_{\pm 6.55}$ & $29.02_{\pm 3.80}$ \\
& Tip-Adapter  & $37.52_{\pm 2.12}$ & $40.98_{\pm 13.52}$ & $47.31_{\pm 6.23}$  & $61.67_{\pm 5.79}$  & $62.67_{\pm 0.97}$ \\
& Tip-Adapter-F  & $37.34_{\pm 15.72}$ & $38.52_{\pm 4.39}$ & $50.44_{\pm 5.30}$ & $43.87_{\pm 2.18}$ & $53.86_{\pm 4.99}$ \\
& Standard LP  & $30.67_{\pm 13.12}$  & $38.13_{\pm 10.28}$  & $49.77_{\pm 8.34}$  & $51.02_{\pm 2.99}$  & $55.34_{\pm 3.56}$ \\
& LP++  & $26.93_{\pm 3.93}$ & $26.16_{\pm 11.70}$ & $36.29_{\pm 9.19}$ & $45.78_{\pm 2.74}$ & $50.98_{\pm 2.14}$ \\
& CoOp  & $25.88_{\pm 9.07}$ & $38.92_{\pm 6.01}$  &  $43.71_{\pm 6.27}$  & $46.8_{\pm 6.80}$  &  $51.07_{\pm 2.56}$ \\
& CoCoOp  & $24.51_{\pm 4.22}$ &  $24.96_{\pm 0.76}$ &  $25.29_{\pm 5.61}$  &  $40.42_{\pm 2.44}$ & $40.97_{\pm 6.50}$  \\
& KgCoOp  & $27.1_{\pm 10.81}$ & $30.28_{\pm 4.45}$ & $35.35_{\pm 8.07}$ & $38.79_{\pm 4.85}$ & $36.59_{\pm 2.32}$ \\
& ProGrad  & $33.98_{\pm 10.76}$ & $37.66_{\pm 6.74}$ & $43.69_{\pm 10.96}$ & $51.07_{\pm 2.47}$ & $46.33_{\pm 5.13}$ \\
\rowcolor{tabhighlight} &  BiomedCoOp (Ours) & $58.64_{\pm 4.71}$ & $57.17_{\pm 1.28}$ & $60.07_{\pm 1.81}$ & $61.98_{\pm 0.77}$ & $62.59_{\pm 1.83}$ \\
\bottomrule
\end{tabular}
}
    \caption{Per-dataset performance comparison of BiomedCoOp with various methods in few-shot setting in terms of classification accuracy (\%).}
    \label{tab_appendix:few_shot_experiments}
\vspace{-5mm}
\end{table*}


\begin{table*}[ht]
\centering
\tablestyle{-5pt}{1.1}
\addtolength{\tabcolsep}{+20pt}
\resizebox{\textwidth}{!}{%
\begin{tabular}{ll|ccccc}
\toprule
\textbf{Dataset} & 
 \textbf{Method} & 
$K$ = 1 &
$K$ = 2 & 
$K$ = 4 & 
$K$ = 8 &
$K$ = 16 \\  \midrule

\multirow{12}{*}{Kvasir}      & BiomedCLIP  & \multicolumn{5}{c}{$54.58$}\\ 
& BiomedCLIP + Ensemble & \multicolumn{5}{c}{$57.5$} \\
& CLIP-Adapter  & $54.83_{\pm 0.48}$ & $54.83_{\pm 0.48}$ & $54.83_{\pm 0.48}$ & $56.08_{\pm 0.86}$ & $56.50_{\pm 1.00}$ \\
& Tip-Adapter  & $56.72_{\pm 3.42}$ & $60.94_{\pm 5.30}$ & $69.61_{\pm 2.06}$  & $69.13_{\pm 1.44}$  & $74.22_{\pm 1.51}$ \\
& Tip-Adapter-F  & $59.19_{\pm 0.89}$ & $64.22_{\pm 3.24}$ & $69.94_{\pm 2.28}$ & $75.86_{\pm 1.00}$ & $78.00_{\pm 1.06}$ \\
& Standard LP  & $54.30_{\pm 2.04}$  & $62.00_{\pm 0.81}$  & $72.38_{\pm 2.65}$  & $78.88_{\pm 0.73}$  & $79.00_{\pm 0.81}$ \\
& LP++  & $58.27_{\pm 3.95}$ & $60.47_{\pm 3.24}$ & $69.36_{\pm 0.84}$ & $72.52_{\pm 2.89}$ & $75.41_{\pm 1.21}$ \\
& CoOp  & $58.2_{\pm 1.64}$ &  $64.86_{\pm 1.4}$ &  $70.78_{\pm 0.31}$  &  $77.14_{\pm 1.25}$ & $77.88_{\pm 0.12}$  \\
& CoCoOp  & $59.45_{\pm 3.25}$ &  $65.5_{\pm 3.41}$ &  $68.94_{\pm 1.29}$  &  $72.92_{\pm 1.46}$ & $75.22_{\pm 2.04}$  \\
& KgCoOp  & $61.67_{\pm 2.16}$ & $65.67_{\pm 1.94}$ & $68.28_{\pm 0.35}$ & $72.05_{\pm 1.8}$ & $72.95_{\pm 1.31}$ \\
& ProGrad  & $60.78_{\pm 0.24}$ & $64.70_{\pm 0.53}$ & $70.00_{\pm 0.24}$ & $76.03_{\pm 1.50}$ & $75.88_{\pm 0.95}$ \\
\rowcolor{tabhighlight} &  BiomedCoOp (Ours) & $62.17_{\pm 1.95}$ & $67.25_{\pm 2.59}$ & $74.08_{\pm 1.10}$ & $77.72_{\pm 0.52}$ & $78.89_{\pm 1.21}$ \\
\midrule
\multirow{12}{*}{CHMNIST}      & BiomedCLIP  & \multicolumn{5}{c}{$30.65$}\\ 
& BiomedCLIP + Ensemble & \multicolumn{5}{c}{$31.52$} \\
& CLIP-Adapter  & $31.27_{\pm 0.69}$ & $31.67_{\pm 0.88}$ & $33.26_{\pm 0.39}$ & $36.48_{\pm 1.32}$ & $42.06_{\pm 2.40}$ \\
& Tip-Adapter  & $46.14_{\pm 9.62}$ & $63.32_{\pm 2.58}$ & $70.05_{\pm 1.11}$  & $69.57_{\pm 1.63}$  & $77.68_{\pm 1.42}$ \\
& Tip-Adapter-F  & $52.81_{\pm 3.10}$ & $58.90_{\pm 4.95}$ & $71.74_{\pm 2.72}$ & $74.51_{\pm 2.43}$ & $80.43_{\pm 2.85}$ \\
& Standard LP  & $58.44_{\pm 2.02}$  & $64.42_{\pm 3.81}$  & $71.07_{\pm 2.23}$  & $76.30_{\pm 3.22}$  & $80.34_{\pm 1.83}$ \\
& LP++  & $57.18_{\pm 6.46}$ & $60.61_{\pm 1.26}$ & $67.79_{\pm 6.97}$ & $72.40_{\pm 0.85}$ & $78.32_{\pm 1.48}$ \\
& CoOp  & $57.34_{\pm 4.2}$ & $59.68_{\pm 1.12}$  &  $68.66_{\pm 2.14}$  & $75.00_{\pm 0.82}$  &  $79.63_{\pm 1.26}$ \\
& CoCoOp  & $49.07_{\pm 4.41}$ &  $50.82_{\pm 3.41}$ &  $58.58_{\pm 2.15}$  &  $66.58_{\pm 1.14}$ & $72.16_{\pm 0.52}$  \\
& KgCoOp  & $59.02_{\pm 4.1}$ & $60.06_{\pm 1.12}$ & $68.77_{\pm 1.02}$ & $69.50_{\pm 0.07}$ & $73.58_{\pm 1.19}$ \\
& ProGrad  & $60.15_{\pm 5.76}$ & $59.60_{\pm 1.53}$ & $69.13_{\pm 1.39}$ & $70.99_{\pm 0.36}$ & $75.11_{\pm 1.50}$ \\
\rowcolor{tabhighlight} &  BiomedCoOp (Ours) & $59.82_{\pm 2.43}$ & $59.79_{\pm 1.36}$ & $71.19_{\pm 1.74}$ & $74.78_{\pm 1.19}$ & $79.05_{\pm 2.24}$ \\
\midrule
\multirow{12}{*}{LC25000}      & BiomedCLIP  & \multicolumn{5}{c}{$50.03$}\\ 
& BiomedCLIP + Ensemble & \multicolumn{5}{c}{$61.84$} \\
& CLIP-Adapter  & $54.83_{\pm 0.36}$ & $53.47_{\pm 2.95}$ & $52.91_{\pm 1.70}$ & $56.33_{\pm 0.45}$ & $57.56_{\pm 1.13}$ \\
& Tip-Adapter  & $75.37_{\pm 4.02}$ & $72.73_{\pm 8.09}$ & $83.32_{\pm 3.95}$  & $87.25_{\pm 1.75}$  & $89.17_{\pm 0.41}$ \\
& Tip-Adapter-F  & $74.21_{\pm 4.35}$ & $71.82_{\pm 7.31}$ & $79.57_{\pm 10.02}$ & $90.41_{\pm 2.43}$ & $92.35_{\pm 1.08}$ \\
& Standard LP  & $74.50_{\pm 2.61}$  & $78.40_{\pm 7.36}$  & $85.30_{\pm 3.56}$  & $90.24_{\pm 0.41}$  & $92.77_{\pm 1.17}$ \\
& LP++  & $63.05_{\pm 9.52}$ & $71.42_{\pm 3.04}$ & $82.61_{\pm 2.31}$ & $89.14_{\pm 2.07}$ & $92.58_{\pm 0.38}$ \\
& CoOp  & $71.90_{\pm 3.53}$ & $76.55_{\pm 2.81}$  &  $84.66_{\pm 2.26}$  & $87.50_{\pm 0.26}$  &  $92.19_{\pm 0.48}$ \\
& CoCoOp  & $63.66_{\pm 4.49}$ &  $71.76_{\pm 0.55}$ &  $77.44_{\pm 2.47}$  &  $85.57_{\pm 1.83}$ & $87.38_{\pm 0.52}$  \\
& KgCoOp  & $71.80_{\pm 2.13}$ & $75.18_{\pm 1.05}$ & $82.10_{\pm 2.35}$ & $84.63_{\pm 0.30}$ & $86.79_{\pm 0.53}$ \\
& ProGrad  & $72.48_{\pm 3.22}$ & $74.76_{\pm 1.40}$ & $84.72_{\pm 2.85}$ & $87.86_{\pm 0.70}$ & $90.70_{\pm 0.66}$ \\
\rowcolor{tabhighlight} &  BiomedCoOp (Ours) & $77.56_{\pm 2.84}$ & $77.74_{\pm 2.00}$ & $85.60_{\pm 1.61}$ & $88.77_{\pm 1.14}$ & $92.68_{\pm 0.57}$ \\
\midrule
\multirow{12}{*}{RETINA}      & BiomedCLIP  & \multicolumn{5}{c}{$26.26$}\\ 
& BiomedCLIP + Ensemble & \multicolumn{5}{c}{$39.27$} \\
& CLIP-Adapter  & $25.49_{\pm 0.46}$ & $25.49_{\pm 0.46}$ & $26.07_{\pm 0.46}$ & $25.84_{\pm 0.87}$ & $26.05_{\pm 0.43}$ \\
& Tip-Adapter  & $26.52_{\pm 0.42}$ & $31.07_{\pm 3.84}$ & $43.42_{\pm 7.04}$  & $48.08_{\pm 7.40}$  & $54.23_{\pm 5.13}$ \\
& Tip-Adapter-F  & $39.53_{\pm 10.83}$ & $33.07_{\pm 5.63}$ & $47.37_{\pm 6.70}$ & $56.07_{\pm 2.57}$ & $62.85_{\pm 1.10}$ \\
& Standard LP  & $39.35_{\pm 6.96}$  & $46.03_{\pm 0.79}$  & $51.31_{\pm 6.52}$  & $53.94_{\pm 1.98}$  & $62.27_{\pm 2.80}$ \\
& LP++  & $35.77_{\pm 5.75}$ & $39.37_{\pm 7.35}$ & $46.95_{\pm 10.07}$ & $53.44_{\pm 1.95}$ & $60.62_{\pm 1.46}$ \\
& CoOp  & $35.02_{\pm 1.40}$ & $35.26_{\pm 3.34}$  &  $42.22_{\pm 3.09}$  & $51.87_{\pm 1.78}$  &  $59.38_{\pm 0.87}$ \\
& CoCoOp  & $32.94_{\pm 0.75}$ &  $36.43_{\pm 4.05}$ &  $39.75_{\pm 3.99}$  &  $48.45_{\pm 1.39}$ & $53.91_{\pm 1.52}$  \\
& KgCoOp  & $33.54_{\pm 2.77}$ & $35.17_{\pm 2.48}$ & $42.61_{\pm 3.16}$ & $49.97_{\pm 2.24}$ & $51.18_{\pm 1.66}$ \\
& ProGrad  & $33.49_{\pm 1.98}$ & $36.49_{\pm 4.64}$ & $43.09_{\pm 3.89}$ & $52.26_{\pm 2.38}$ & $50.47_{\pm 2.40}$ \\
\rowcolor{tabhighlight} &  BiomedCoOp (Ours) & $36.64_{\pm 3.34}$ & $38.67_{\pm 1.79}$ & $45.58_{\pm 5.03}$ & $56.47_{\pm 1.37}$ & $61.28_{\pm 1.06}$ \\
\midrule
\multirow{12}{*}{KneeXray}      & BiomedCLIP  & \multicolumn{5}{c}{$29.53$}\\ 
& BiomedCLIP + Ensemble & \multicolumn{5}{c}{$39.37$} \\
& CLIP-Adapter  & $29.00_{\pm 0.17}$ & $28.66_{\pm 0.45}$ & $28.96_{\pm 0.46}$ & $28.80_{\pm 0.20}$ & $29.08_{\pm 0.32}$ \\
& Tip-Adapter  & $29.04_{\pm 0.67}$ & $33.55_{\pm 5.96}$ & $24.19_{\pm 4.23}$  & $25.76_{\pm 3.35}$  & $33.17_{\pm 7.59}$ \\
& Tip-Adapter-F  & $30.01_{\pm 0.50}$ & $28.38_{\pm 2.18}$ & $26.59_{\pm 5.70}$ & $26.46_{\pm 2.20}$ & $27.67_{\pm 3.21}$ \\
& Standard LP  & $26.02_{\pm 11.08}$  & $26.57_{\pm 5.17}$  & $27.83_{\pm 4.92}$  & $22.20_{\pm 3.68}$  & $23.97_{\pm 3.55}$ \\
& LP++  & $21.25_{\pm 8.60}$ & $26.40_{\pm 3.26}$ & $28.92_{\pm 4.97}$ & $23.75_{\pm 2.50}$ & $26.38_{\pm 3.39}$ \\
& CoOp  & $24.96_{\pm 9.41}$ & $25.89_{\pm 5.06}$  &  $23.85_{\pm 4.25}$  & $26.23_{\pm 4.01}$  &  $28.48_{\pm 1.84}$ \\
& CoCoOp  & $25.42_{\pm 6.38}$ &  $28.85_{\pm 8.24}$ &  $30.66_{\pm 4.49}$  &  $21.78_{\pm 8.29}$ & $24.86_{\pm 4.15}$  \\
& KgCoOp  & $29.07_{\pm 3.31}$ & $28.14_{\pm 4.53}$ & $22.44_{\pm 2.88}$ & $23.37_{\pm 3.35}$ & $24.80_{\pm 0.47}$ \\
& ProGrad  & $30.09_{\pm 6.00}$ & $23.83_{\pm 0.57}$ & $23.95_{\pm 2.87}$ & $24.78_{\pm 2.32}$ & $26.27_{\pm 3.29}$ \\
\rowcolor{tabhighlight} &  BiomedCoOp (Ours) & $36.13_{\pm 1.75}$ & $37.72_{\pm 0.54}$ & $35.91_{\pm 0.54}$ & $37.7_{\pm 1.00}$ & $39.69_{\pm 1.75}$ \\
\bottomrule
\end{tabular}
}
    \caption*{Table \ref*{tab_appendix:few_shot_experiments} (continued): Per-dataset performance comparison of BiomedCoOp with various methods in few-shot setting in terms of classification accuracy (\%).}
\vspace{-5mm}
\end{table*}


\begin{table*}[ht]
\centering
\tablestyle{-5pt}{1.1}
\addtolength{\tabcolsep}{+20pt}
\resizebox{\textwidth}{!}{%
\begin{tabular}{ll|ccccc}
\toprule
\textbf{Dataset} & 
 \textbf{Method} & 
$K$ = 1 &
$K$ = 2 & 
$K$ = 4 & 
$K$ = 8 &
$K$ = 16 \\  \midrule
\multirow{12}{*}{OCTMNIST}      & BiomedCLIP  & \multicolumn{5}{c}{$30.00$}\\ 
& BiomedCLIP + Ensemble & \multicolumn{5}{c}{$47.40$} \\
& CLIP-Adapter  & $44.00_{\pm 5.79}$ & $49.73_{\pm 2.41}$ & $49.96_{\pm 1.77}$ & $49.50_{\pm 3.33}$ & $52.73_{\pm 0.62}$ \\
& Tip-Adapter  & $32.36_{\pm 3.94}$ & $33.8_{\pm 6.16}$ & $38.10_{\pm 5.01}$  & $53.93_{\pm 3.17}$  & $53.33_{\pm 3.92}$ \\
& Tip-Adapter-F  & $46.66_{\pm 2.58}$ & $53.93_{\pm 1.67}$ & $55.20_{\pm 4.75}$ & $65.00_{\pm 6.61}$ & $72.50_{\pm 1.38}$ \\
& Standard LP  & $47.25_{\pm 12.64}$  & $54.21_{\pm 8.23}$  & $61.00_{\pm 7.07}$  & $65.85_{\pm 9.01}$  & $69.40_{\pm 3.68}$ \\
& LP++  & $47.24_{\pm 13.84}$ & $53.18_{\pm 9.08}$ & $59.02_{\pm 8.59}$ & $63.69_{\pm 8.26}$ & $68.35_{\pm 7.42}$ \\
& CoOp  & $52.63_{\pm 2.95}$ & $53.57_{\pm 3.86}$  &  $53.37_{\pm 2.35}$  & $63.67_{\pm 4.47}$  &  $65.47_{\pm 7.47}$ \\
& CoCoOp  & $49.33_{\pm 4.58}$ &  $50.93_{\pm 8.01}$ &  $48.57_{\pm 6.25}$  &  $55.40_{\pm 1.88}$ & $60.67_{\pm 3.41}$  \\
& KgCoOp  & $50.63_{\pm 3.18}$ & $50.53_{\pm 5.39}$ & $52.97_{\pm 1.58}$ & $61.03_{\pm 3.78}$ & $62.80_{\pm 3.85}$ \\
& ProGrad  & $51.40_{\pm 3.05}$ & $55.33_{\pm 3.38}$ & $55.07_{\pm 1.22}$ & $62.17_{\pm 6.01}$ & $63.33_{\pm 6.15}$ \\
\rowcolor{tabhighlight} &  BiomedCoOp (Ours) & $51.83_{\pm 1.52}$ & $55.03_{\pm 4.72}$ & $54.73_{\pm 1.86}$ & $58.87_{\pm 5.35}$ & $66.93_{\pm 2.13}$ \\
\midrule
\multirow{12}{*}{\textbf{Average}}      & BiomedCLIP  & \multicolumn{5}{c}{$42.05$}\\ 
& BiomedCLIP + Ensemble & \multicolumn{5}{c}{$52.27$} \\
& CLIP-Adapter  & $44.66_{\pm 2.97}$ & $43.91_{\pm 2.48}$ & $44.36_{\pm 1.94}$ & $45.42_{\pm 2.38}$ & $46.69_{\pm 1.71}$ \\
& Tip-Adapter  & $49.19_{\pm 4.84}$ & $52.36_{\pm 6.57}$ & $57.33_{\pm 5.07}$  & $61.98_{\pm 5.76}$  & $67.15_{\pm 4.25}$ \\
& Tip-Adapter-F  & $51.17_{\pm 8.33}$ & $52.74_{\pm 5.88}$ & $61.23_{\pm 6.22}$ & $65.91_{\pm 3.64}$ & $70.91_{\pm 2.65}$ \\
& Standard LP  & $47.25_{\pm 8.65}$  & $54.21_{\pm 7.80}$  & $61.00_{\pm 6.81}$  & $65.85_{\pm 4.89}$  & $69.40_{\pm 2.91}$ \\
& LP++  & $47.24_{\pm 7.68}$ & $53.18_{\pm 7.29}$ & $59.02_{\pm 6.93}$ & $63.69_{\pm 4.68}$ & $68.35_{\pm 3.59}$ \\
& CoOp  & $50.16_{\pm 6.93}$ & $54.18_{\pm 4.31}$  &  $59.75_{\pm 3.72}$  & $65.84_{\pm 3.66}$  &  $69.62_{\pm 2.83}$ \\
& CoCoOp  & $48.49_{\pm 4.39}$ &  $51.28_{\pm 5.06}$ &  $54.69_{\pm 4.79}$  &  $61.08_{\pm 3.49}$ & $65.09_{\pm 2.87}$  \\
& KgCoOp  & $51.83_{\pm 5.53}$ & $53.47_{\pm 5.07}$ & $58.59_{\pm 4.50}$ & $63.65_{\pm 2.73}$ & $64.88_{\pm 1.95}$ \\
& ProGrad  & $51.88_{\pm 6.39}$ & $54.71_{\pm 4.46}$ & $60.42_{\pm 4.78}$ & $65.61_{\pm 3.02}$ & $67.13_{\pm 3.00}$ \\
\rowcolor{tabhighlight} &  BiomedCoOp (Ours) & $57.03_{\pm 2.80}$ & $59.13_{\pm 3.64}$ & $63.95_{\pm 2.42}$ & $68.32_{\pm 2.65}$ & $72.42_{\pm 1.62}$ \\
\bottomrule
\end{tabular}
}
    \caption*{Table \ref*{tab_appendix:few_shot_experiments} (continued): Per-dataset performance comparison of BiomedCoOp with various methods in few-shot setting in terms of classification accuracy (\%).}
\vspace{-5mm}
\end{table*}

%% file: tables/interpret_prompts.tex
\begin{table*}
\tablestyle{-26pt}{1.1}
\addtolength{\tabcolsep}{+30pt}
\resizebox{\textwidth}{!}{%
\begin{tabular}{c cccc}
\toprule
Dataset      & Context Token \#1                         & Context Token \#2                        & Context Token \#3                     & Context Token \#4                       \\ \midrule
BTMRI    & \texttt{mri}   (2.4971)    & \texttt{curcumin} (2.5835)           & \texttt{of} (1.5667)   & \texttt{a} (1.6353) \\ \midrule
BUSI    & \texttt{a}   (2.5550)    & \texttt{photo} (3.5649)           & \texttt{of} (2.1298)   & \texttt{b} (3.4897) \\ \midrule
COVID-QU-Ex     & \texttt{measured}   (2.1999)        & \texttt{image} (2.2856)          & \texttt{of}  (1.9166)       & \texttt{a} (1.9205)     \\ \midrule
CTKIDNEY    & \texttt{a}   (2.1290)    & \texttt{schem} (2.6564)           & \texttt{right} (2.3790)        & \texttt{a} (1.7574)   \\ \midrule
DermaMNIST    & \texttt{dextrose} (2.8292)      & \texttt{photo} (3.1084)        & \texttt{ricin} (3.2378)         & \texttt{autologous} (3.0297)      \\ \midrule
Kvasir & \texttt{endoscopy}  (2.1880)      & \texttt{scar} (2.4835)     & \texttt{of} (2.2698)                  & \texttt{maintained}  (2.4771)     \\ \midrule
CHMNIST     & \texttt{a}   (3.0301)            & \texttt{original} (3.4248)    & \texttt{composed} (2.2125)         & \texttt{discern} (3.4506)         \\ \midrule
LC25000  & \texttt{a}   (1.5298)      & \texttt{photo} (2.3540)            & \texttt{of} (1.6363) & \texttt{a} (2.0292)     \\ \midrule
RETINA & \texttt{a}   (1.5986)     & \texttt{papill} (2.3636)        & \texttt{of}  (1.6976) & \texttt{receptive} (2.1135)      \\ \midrule
KneeXray   & \texttt{a }  (4.2063)       & \texttt{calcification} (5.4999)          & \texttt{osteoc} (2.8673)  & \texttt{showed} (2.9774)        \\ \midrule
OCTMNIST  & \texttt{localized} (2.1744) & \texttt{example} (3.6750)          & \texttt{of} (1.8752)     & \texttt{possible} (2.4803)    \\ \bottomrule
\end{tabular}}
\caption{The nearest words for each of the 4 context vectors learned by BiomedCoOp, with their distances to the corresponding context tokens shown in parentheses.}
\label{tab:interpret_prompts}
\end{table*}